\documentclass[11pt]{article}

\usepackage[preprint]{acl}

\usepackage{times}
\usepackage{latexsym}
\usepackage[T1]{fontenc}
\usepackage[utf8]{inputenc}
\usepackage{microtype}
\usepackage{graphicx}
\usepackage{booktabs}
\usepackage{amsfonts}
\usepackage{amsmath}
\usepackage{amssymb}
\usepackage{algorithm}
\usepackage{algorithmic}
\usepackage{multirow}
\usepackage{subcaption}
\usepackage{float}
\usepackage{tikz}
\usetikzlibrary{calc}
\usepackage{pgfplots}
\pgfplotsset{compat=1.17}
\usepgfplotslibrary{groupplots}
\usepackage{amsthm}
\usepackage{listings}
\usepackage[most]{tcolorbox}
\definecolor{promptbg}{HTML}{EEF2F7}
\definecolor{promptframe}{HTML}{3B6CB0}
\definecolor{metricbg}{HTML}{F0F4FA}
\definecolor{metricframe}{HTML}{5B7FA6}
\definecolor{hpbg}{HTML}{F2F7F2}
\definecolor{hpframe}{HTML}{4A8C6F}
\newtheorem{definition}{Definition}

\newcommand{\ours}{\textsc{LLMZero}}
\newcommand{\gain}[1]{\textsubscript{\scriptsize(+#1)}}

\newcommand{\revised}[1]{{\color{black}#1}}

\title{\ours{}: Discovering Adaptive Training Strategies\\for RL Post-Training via LLM Agents}

\author{
  Haoyang Fang$^{\dagger}$, 
  Wei Zhu$^{\dagger}$$^{*}$, 
  Boran Han$^{\dagger}$, 
  Alex Zhang, 
  \\
  \textbf{Zhenyu Pan$^{*}$}, 
  \textbf{Shuo Yang$^{*}$}, 
  \textbf{Shuai Zhang$^{*}$}, 
  \textbf{Jiading Gai$^{*}$}, 
  \textbf{Peng Tang}, \\
  \textbf{Cuixiong Hu$^{*}$}, 
  \textbf{Xuan Zhu$^{*}$}, 
  \textbf{Huzefa Rangwala$^{*}$}, 
  \textbf{George Karypis$^{*}$}, 
  \textbf{Bernie Wang}$^{\dagger}$$^{*}$ \\
  Amazon \\
  \texttt{\{haoyfang, weizhuq, boranhan, yuyawang\}@amazon.com}
}

\newcommand\blfootnote[1]{%
  \begingroup
  \renewcommand\thefootnote{}\footnote{#1}%
  \addtocounter{footnote}{-1}%
  \endgroup
}

\begin{document}

\maketitle

\blfootnote{$^{\dagger}$LLMZero Project Core Team.}
\blfootnote{$^{*}$Work done at Amazon.}

\begin{abstract}
RL post-training strategies are dataset-dependent and reveal a recurring empirical pattern: capacity parameters accumulate monotonically across stages, while regularization parameters predominantly oscillate in response to shifting training dynamics. This distinction highlights a potential flaw in fixed training schedules: by forcing all parameters along rigid paths, they fail to capture the dynamic exploration-exploitation tradeoffs that regularization must track. We uncover this through \ours{}, an agentic system that optimizes training trajectories via tree search by diagnosing pathologies at each checkpoint and proposing coordinated multi-parameter transitions. Across four diverse GRPO tasks, \ours{} discovers strategies that improve over the base model by 9\% to 140\% and over grid search by 6\% to 15\% (relative), consistently outperforming random search and a skill-based agent \revised{under a matched compute budget}. \revised{The capacity--regularization asymmetry is consistent across all four tasks, offering a candidate design heuristic for multi-stage training.}
\end{abstract}

\section{Introduction}
\label{sec:intro}

Fixed training schedules are suboptimal for RL post-training~\cite{unifypt,dump}. In most recent works, the community has converged on a narrow set of progressive scheduling techniques with all other hyperparameters held constant, applied identically regardless of dataset, model size, or emergent training dynamics. The dominant approach is gradually increasing response length~\citep{deepscaler,acereason,skywork_or1,jt_math,mimo,deepcoder,still3,p1vl,diff_aware_staged_rl}. Others gradually increase rollouts~\citep{deepscaler,acereason,fastcurl,p1vl}, stage training data by progressive difficulty~\citep{acereason,fastcurl,skywork_or1,tacler,qwenlong_l1,p1vl,diff_aware_staged_rl}, or adopt an oscillating response length schedule~\citep{fastcurl}. This practice is motivated by training base models to produce increasingly long chains of thought, but is less well-justified for continued training on models that already generate extended reasoning. These guidebook-driven schedules do not systematically specify \emph{when} to trigger a transition, \emph{how much} to adjust, or \emph{which} parameters to change across different tasks. When training dynamics deviate from expectations (KL divergence spikes, model collapse, stagnating validation), no systematic mechanism responds.

To address this, we propose \ours{}\footnote{\revised{Code will be released at: \url{https://github.com/amazon-science/LLMZero}. Until then, please contact haoyfang@amazon.com.}}, an agent-driven system that searches over training trajectories. The strategies \ours{} discovers reveal a recurring structural asymmetry: \emph{capacity parameters (response length, rollouts) accumulate monotonically across all four tasks, while regularization parameters (learning rate, KL coefficient, temperature) predominantly oscillate}. \revised{Capacity parameters are information-constructive, e.g., shortening response length truncates learned generations and raises gradient variance.} Regularization parameters track a non-stationary tradeoff where the optimal exploration-exploitation balance shifts continuously during training, making monotonic decay a poor fit in practice. This \revised{pattern} manifests differently per task (ChemCoTBench~\citep{chemcotbench} uses 5-stage progressive stabilization with reactive KL spikes, SSMR-Bench~\citep{ssmrbench} benefits from LR/KL oscillation with monotonic capacity expansion, and PaperSearchQA~\citep{papersearchqa} uses a ``tighten then loosen'' pattern to escape convergence plateaus), but the underlying asymmetry between parameter classes is consistent \revised{across our four evaluated tasks} (\S\ref{sec:strategy_analysis}).

Traditional adaptive controllers (e.g., proportional KL adjustment~\citep{ppo}) fall short in this setting because they tune only one parameter based on one signal. The strategies we discover require \emph{coordinated} multi-dimensional transitions \revised{that change 3+ parameters at once}, such as simultaneously raising learning rate to escape a plateau while increasing KL penalty to prevent larger steps from causing divergence. These coordinated interventions require understanding the causal relationships between parameters and training dynamics, which is what LLM reasoning provides \revised{(Appendix~\ref{app:reasoning_trace} gives a full trace)}.

\ours{} builds a tree of training trajectories where LLM agents analyze training dynamics through textual metrics and visual plots, and then propose targeted hyperparameter transitions conditioned on the observed training state. An agentic early stopper terminates unpromising branches in real time, reallocating compute toward more promising branches. A modified Monte Carlo Tree Search (MCTS) balances deepening promising branches against exploring alternatives, while checkpoint-based branching enables multi-stage strategies (\S\ref{sec:search}).

Across ChemCoTBench, PaperSearchQA, SSMR-Bench, and WildSci~\citep{wildsci}, \ours{} \revised{achieves relative gains of 9--140\% over the base model and 6--15\% over grid search, and surpasses random search and the skill-based agent} \revised{under a matched compute budget (\S\ref{sec:experiments}; Appendix~\ref{app:budget_fairness})}. Notably, \ours{} finds its best strategy within the first 12 iterations on 3 of 4 tasks, demonstrating its high search efficiency.

Beyond the system itself, our findings demonstrate that optimal strategies are dataset-dependent and not well captured by fixed guidebooks (\S\ref{sec:patterns}). Furthermore, \ours{} consistently improves over base configurations from 0.6B to 8B parameters, indicating that dynamics-aware strategy search generalizes robustly across model scales (\S\ref{sec:scaling}).

\begin{figure*}[t]
    \centering
    \includegraphics[width=\textwidth]{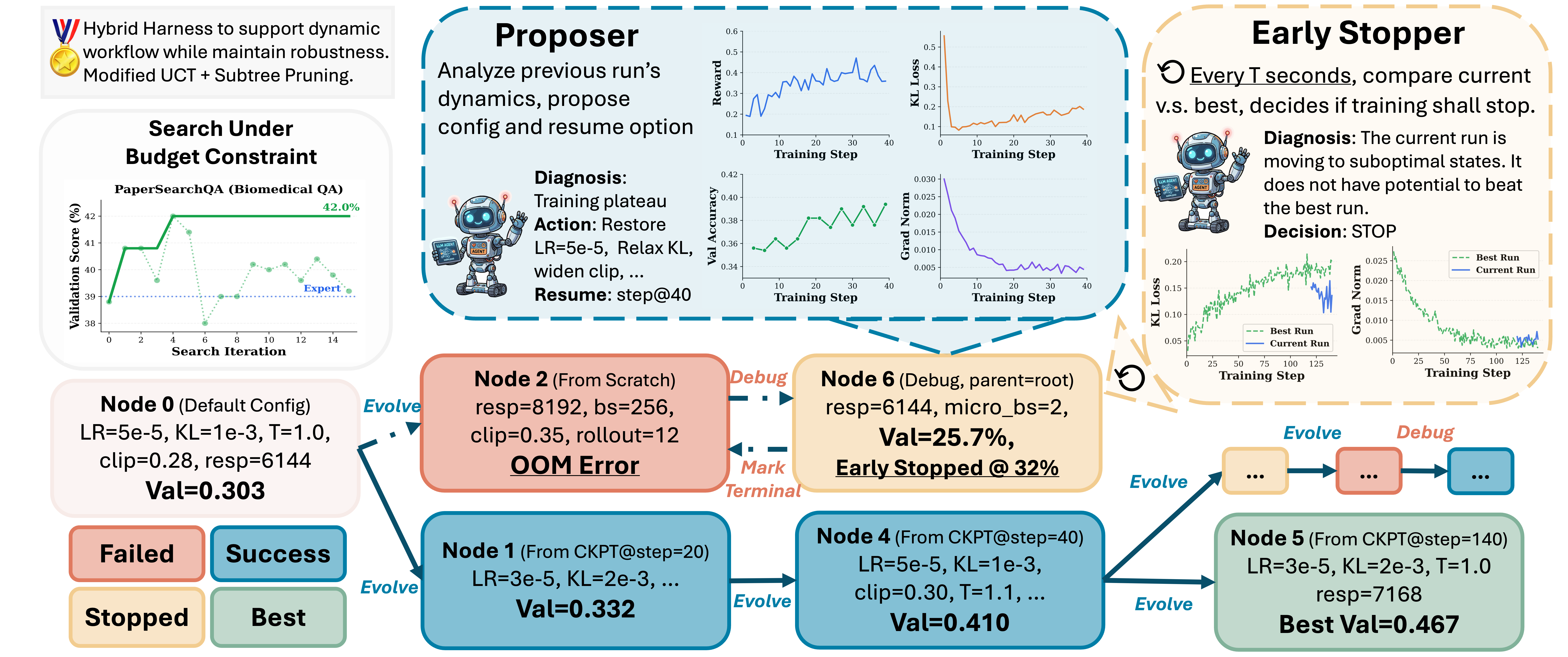}
    \caption{Overview of \ours{}. The system builds a tree of training trajectories where each node stores a full hyperparameter configuration and resumes from a parent checkpoint, composing multi-stage adaptive strategies via backtracking. At each iteration, the \emph{proposer agent} analyzes training dynamics (rewards, KL divergence, validation scores, gradient norms) through both text summaries and visual plots, then proposes a new configuration with a checkpoint to resume from. During training, the \emph{early stopper} periodically overlays the current run's trajectory against the best completed strategy and terminates dominated runs. Metric values and scores are purely illustrative and do not reflect an actual training run.}
    \label{fig:overview}
\end{figure*}

\section{Preliminary}
\label{sec:preliminary}

\subsection{Training Strategy Formalization}
\label{sec:hierarchy}

We formalize three paradigms of increasing complexity for RL post-training. Let $M_0$ denote the base model, $\Theta$ the hyperparameter space, $\mathcal{H}_t = \{(s_1, r_1), \ldots, (s_t, r_t)\}$ the training history up to step $t$, and $\mu$ a validation metric.

\begin{definition}[Single-Stage Training]
\label{def:single_stage}
A single-stage strategy selects one fixed configuration and trains to completion:
\begin{equation}
    \sigma_{\text{static}} = \langle (\theta, 0) \rangle, \quad \theta^* = \arg\max_{\theta \in \Theta}\; \mu\!\left(\mathcal{T}(M_0, \theta)\right).
    \label{eq:single_stage}
\end{equation}
HPO methods (grid, random, Bayesian) search over $\Theta$ by running multiple independent static trials.
\end{definition}

\begin{definition}[Multi-Stage Training]
\label{def:multi_stage}
A multi-stage strategy is a \emph{guidebook-driven} sequence of $L > 1$ phases:
\begin{align}
    \sigma_{\text{multi}} &= \langle (\theta_1, k_1), (\theta_2, k_2), \ldots, (\theta_L, k_L) \rangle, \nonumber \\
    &\quad \theta_\ell \in \Theta, \; k_\ell \in \mathbb{N},
    \label{eq:multi_stage}
\end{align}
where phase $\ell$ trains with configuration $\theta_\ell$ starting from step $k_\ell$. The schedule structure is specified before training begins and does not systematically depend on training history $\mathcal{H}_t$. 
\end{definition}

\begin{definition}[Adaptive Training]
\label{def:adaptive}
An adaptive strategy selects both the configuration and the checkpoint to resume from based on observations from prior phases. A transition policy $\pi$ selects:
\begin{equation}
    (\theta_\ell,\, k_\ell,\, j_\ell) = \pi\!\left(\{(\theta_i, k_i, j_i, \mathcal{H}_i)\}_{i < \ell}\right),
    \label{eq:adaptive}
\end{equation}
where $j_\ell \in \{1, \ldots, \ell{-}1\}$ identifies which prior phase to resume from. The policy can backtrack to any earlier checkpoint, enabling branching. Neither the number of phases, configurations, transition points, nor resumption targets are determined before training begins.
\end{definition}

RL training is inherently non-stationary: the pace at which exploration must yield to exploitation depends on the dataset, model size, and reward structure, all of which are difficult to predict before training begins. An adaptive strategy can respond in real time, but the space of possible transition policies is vast, motivating automated search.

\section{\ours{}}
\label{sec:method}

\ours{} (Figure~\ref{fig:overview}) builds a tree of training trajectories where each branch point represents a hyperparameter transition chosen based on observed training dynamics. This section describes how the system discovers adaptive strategies.

\subsection{Problem Formulation}
\label{sec:formulation}

Given a dataset $\mathcal{D} = \{(x_i, m_i)\}_{i=1}^{N}$, a base model $M_0$, a training procedure $\mathcal{T}$, and a validation metric $\mu: \mathcal{Y} \times \mathcal{M} \to [0,1]$, we seek an adaptive strategy $\sigma^*$ maximizing held-out performance:
\begin{align}
    \sigma^* &= \operatorname*{arg\,max}_{\sigma}\, \mathbb{E}_{(x,m) \sim \mathcal{D}_{\text{val}}} \!\left[ \mu\!\big(\mathcal{T}(M_0, \sigma)(x), m\big) \right]\!, \nonumber \\
    &\quad \text{s.t.} \quad \text{\#iterations} \leq B,
    \label{eq:objective}
\end{align}
where $\sigma = \langle (\theta_1, k_1), \ldots, (\theta_L, k_L) \rangle$ is constructed online (\S\ref{sec:hierarchy}) under a small budget $B$ (typically 4--16 iterations, each requiring hours of GPU time).

We model the search as a tree problem. Each node represents one training phase. The root uses a default configuration. Children are created by resuming from a parent checkpoint with modified hyperparameters (\emph{evolving}), by fixing failed runs (\emph{debugging}), or by starting fresh to maintain diversity. Each scratch-to-leaf path forms a candidate multi-stage strategy, and siblings reuse the same parent checkpoint for compute sharing.

\subsection{Tree Search and Subtree Pruning}
\label{sec:search}

\ours{} performs a modified MCTS over training trajectories. Each iteration selects a node via UCT, expands it by proposing a hyperparameter transition (or debugging a failure), executes the training phase, and backpropagates the validation score. We adopt a UCT variant~\cite{exts} with scale-invariant scoring and virtual child competition; details are reproduced in Appendix~\ref{app:mcts} for completeness.

\paragraph{Subtree pruning.} A node is marked \emph{terminal} when it can no longer produce children, and terminal subtrees are excluded from selection. When a failed run is fixed, the successful descendant is reparented as a sibling of the oldest ancestor in the debug chain, and the entire subtree below is pruned. Terminality propagates upward: a node becomes terminal when fully expanded with all children terminal.

\paragraph{The search loop.} Algorithm~\ref{alg:mcts} (Appendix~\ref{app:mcts}) details the complete procedure. The key mechanisms are: (1)~a \emph{proposer agent} that performs multimodal analysis of training dynamics (\S\ref{sec:reflection}), (2)~an \emph{agentic early stopper} that terminates unpromising runs in real time (\S\ref{sec:early_stopping}), and (3)~forced from-scratch injection to maintain diversity (\S\ref{sec:checkpoint}).

\subsection{Checkpoint-Based Strategy Composition}
\label{sec:checkpoint}

When an evolve node is created, it loads its parent's model weights at step $k$ and continues training with modified hyperparameters. Successive transitions compose into a multi-stage strategy along the scratch-to-leaf path.

We resume only the \textbf{model weights}, reinitializing optimizer state and dataloader position. This allows arbitrary configuration changes at each transition (batch size, learning rate, optimizer type) while avoiding inheritance of suboptimal momentum accumulators. Because each checkpoint must persist on disk for potential future resumption, storage cost grows linearly with tree depth and training steps. We use LoRA throughout and save only the adapter weights at each checkpoint. This design choice means the LoRA rank is fixed across all phases of a strategy and cannot be modified at transitions.

\paragraph{Forced from-scratch injection.} To prevent the search from exclusively exploiting a potentially suboptimal initial configuration, we enforce a minimum from-scratch ratio $\rho_{\min}=0.2$. When $n_{\text{scratch}} / n_{\text{evolve}} < \rho_{\min}$, we select the best-scoring node $n^*$, create a from-scratch child that does not count against $n^*$'s branching limit, and withhold checkpoint information from the proposer to ensure a fresh strategy. This is the primary mechanism by which \ours{} escapes a poor default configuration. Otherwise, a bad initial run would anchor all subsequent evolve nodes to a weak checkpoint. 
\revised{This mechanism serves as a robustness safeguard rather than a primary driver of performance: on all four tasks the best strategy originates from the root subtree, so the injection mechanism was not triggered in these primary evaluations (Appendix~\ref{app:budget_fairness}); its value surfaces only under a poor default, where it lifts performance from 51.5\% to 56.1\% ($\mathrm{KL}{=}10^{-2}$; Appendix~\ref{app:from_scratch}).}

\subsection{Dynamics-Aware Transition Proposal}
\label{sec:reflection}

The proposer agent is the core adaptive component. It receives the parent node's training configuration, text summaries of step-level metrics, and the best validation score. It also performs \emph{visual reasoning} over per-metric training curve plots, enabling pattern recognition such as trend inflections, divergence onset, and plateau detection that textual summaries alone may miss. Its reasoning proceeds in four stages: (1)~\emph{diagnose} the parent run's training health from primary metrics and diagnostic signals; (2)~propose \emph{coordinated hyperparameter changes} that address the diagnosed issue, reasoning about causal dependencies between parameters (e.g., raising LR to escape a plateau while increasing KL penalty to prevent divergence); (3)~select a \emph{resumption checkpoint} (either a specific step or from scratch); and (4)~compute the \emph{epoch budget} based on the chosen batch size and checkpoint step. To mitigate common LLM hallucinations or misinterpretations of domain-specific GRPO/PPO metrics, we inject human-written descriptions to ground the agent's reasoning without restricting which parameters it can modify (Appendix~\ref{app:knowledge}). The full prompt template is in Appendix~\ref{app:proposer_prompt}.

\subsection{Agentic Early Stopping}
\label{sec:early_stopping}

At 15-minute intervals during training, the early stopper samples current metrics and generates overlay plots comparing the current run's trajectory against the best completed strategy. It outputs \texttt{CONTINUE} or \texttt{STOP} with 
explicit reasoning about whether the current trajectory is likely to surpass the incumbent's performance (full prompt in Appendix~\ref{app:early_stopper_prompt}). Across all 4 tasks, the early stopper terminated 62.1\% of nodes before completion, reducing total GPU consumption by an estimated 40--60\% relative to running all nodes to completion.

\subsection{Automated Pipeline}
\label{sec:system}

\ours{} includes an automated pipeline that handles data preparation, reward function implementation, training code generation, and job execution. In our evaluation, data processing and reward functions are fixed across all methods to ensure a fair comparison. The adaptive strategy search follows a fixed workflow; the essential LLM-based components are the \textbf{proposer agent} (\S\ref{sec:reflection}) and the \textbf{agentic early stopper} (\S\ref{sec:early_stopping}). Pipeline details are in Appendix~\ref{app:setup}.

\section{Experiments}
\label{sec:experiments}

We evaluate \ours{} across 4 diverse GRPO tasks to answer: \textbf{(RQ1)} Can adaptive strategies outperform static configurations? \textbf{(RQ2)} Are discovered patterns dataset-dependent? \textbf{(RQ3)} How does performance scale with model size? \textbf{(RQ4)} How do individual components contribute? \textbf{(RQ5)} Do strategies transfer across tasks?

\begin{table*}[t]
\caption{Main results: test score (\%) at the best validation node within 16 iterations on Qwen3-4B. \textbf{Bold}: best per column (aggregate only). \underline{Underline}: second best. Subscripts show gain over the base model. ChemCoT reports category-averaged test accuracy across 3 task families (mol.\ optimization, 6 subtasks at 0\% for all methods, is omitted). SSMR shows per-subtask test scores (Scl=scale, Bea=beat, Cho=chord, Int=interval). WildSci practitioner config scores below Qwen3-4B on weighted test aggregate due to domain-level trade-offs (see \S\ref{sec:per_dataset}). }
\label{tab:main_results}
\centering
\resizebox{\textwidth}{!}{%
\setlength{\tabcolsep}{4pt}
\small
\begin{tabular}{ll*{3}{c}c c *{4}{c}c c}
\toprule
& & \multicolumn{4}{c}{\textbf{ChemCoT} {\scriptsize (Chem)}} & \textbf{PaperSearchQA} & \multicolumn{5}{c}{\textbf{SSMR} {\scriptsize (Music)}} & \textbf{WildSci} \\
\cmidrule(lr){3-6} \cmidrule(lr){8-12}
 & \textbf{Type} & {\scriptsize Und} & {\scriptsize Edit} & {\scriptsize Rxn} & {\scriptsize Avg} & {\scriptsize Bio} & {\scriptsize Scl} & {\scriptsize Bea} & {\scriptsize Cho} & {\scriptsize Int} & {\scriptsize Avg} & {\scriptsize Sci} \\
\midrule
Qwen3-4B & -- & 40.5 & 8.0 & 2.3 & 16.9 & 31.6 & 60.0 & 44.8 & 50.4 & 50.4 & 51.4 & 53.6 \\
\midrule
Practitioner & Static & 61.1\gain{20.6} & 23.0\gain{15.0} & 14.9\gain{12.6} & 33.0\gain{16.1} & 39.0\gain{7.4} & 76.8\gain{16.8} & 60.0\gain{15.2} & 65.6\gain{15.2} & 60.8\gain{10.4} & 65.8\gain{14.4} & 53.2\textsubscript{\scriptsize($-$0.4)} \\
Random search & Static & 66.4\gain{25.9} & 23.0\gain{15.0} & 28.7\gain{26.4} & \underline{39.4}\gain{22.5} & 37.6\gain{6.0} & 90.4\gain{30.4} & 77.6\gain{32.8} & 68.8\gain{18.4} & 60.8\gain{10.4} & 74.4\gain{23.0} & 55.8\gain{2.2} \\
Grid search & Static & 61.2\gain{20.7} & 23.4\gain{15.4} & 20.9\gain{18.6} & 35.2\gain{18.3} & 39.0\gain{7.4} & 87.2\gain{27.2} & 84.0\gain{39.2} & 69.6\gain{19.2} & 69.6\gain{19.2} & 77.6\gain{26.2} & 53.0\textsubscript{\scriptsize($-$0.6)} \\
\midrule
Skill-based LLM agent & Adaptive & 61.8\gain{21.3} & 21.4\gain{13.4} & 27.7\gain{25.4} & 37.0\gain{20.1} & \underline{40.2}\gain{8.6} & 96.0\gain{36.0} & 88.0\gain{43.2} & 74.4\gain{24.0} & 61.6\gain{11.2} & \underline{80.0}\gain{28.6} & \underline{56.6}\gain{3.0} \\
\textbf{\ours{}} & \textbf{Adaptive} & 69.8\gain{29.3} & 33.3\gain{25.3} & 18.5\gain{16.2} & \textbf{40.5}\gain{23.6} & \textbf{42.6}\gain{11.0} & 94.4\gain{34.4} & 81.6\gain{36.8} & 77.6\gain{27.2} & 75.2\gain{24.8} & \textbf{82.2}\gain{30.8} & \textbf{58.5}\gain{4.9} \\
\bottomrule
\end{tabular}%
}
\end{table*}

\subsection{Setup}
\label{sec:setup}

\paragraph{Tasks.} We evaluate on ChemCoTBench~\citep{chemcotbench}, PaperSearchQA~\citep{papersearchqa}, SSMR-Bench~\citep{ssmrbench}, and WildSci~\citep{wildsci}. Each dataset is uniformly subsampled to 5{,}000 train, 500 validation, and 500 test examples to keep per-iteration training time tractable for search. All tasks use GRPO via VeRL~\citep{verl} (Appendix~\ref{app:datasets}).

\paragraph{Models.} Primary evaluation uses \textbf{Qwen3-4B} (LoRA, rank=64). Scaling analysis spans \textbf{Qwen3-0.6B} through \textbf{8B}. Base model evaluation: greedy decoding, max response length=8192. Infrastructure: Ray clusters on EKS, 32--64 A100 40G GPUs.

\paragraph{Baselines.} We compare against: (1)~a \textbf{practitioner baseline} (fixed GRPO recipe tuned on separate tasks), (2)~\textbf{random search} (8 trials from broad HP ranges), (3)~\textbf{grid search} (8 trials over LR $\times$ LoRA rank, selected as the most efficient range based on internal experience), and (4)~a \textbf{skill-based LLM agent} built on Claude Code as the orchestration backend with Claude Opus 4.6 at high reasoning effort, which autonomously plans iterations and can stop/resume from any checkpoint without tree search or visual reasoning. Because its autonomous orchestration maintains full conversation context, it costs \revised{73$\times$ more in API overall} than \ours{} with its fixed workflow, limiting it to 6--9 iterations in practice. Full baseline configurations are in Appendix~\ref{app:setup}; budget fairness details are in Appendix~\ref{app:budget_fairness}.

\subsection{Main Results (RQ1)}
\label{sec:main_results}

Table~\ref{tab:main_results} presents the test score at the best validation step for each method. The comparison isolates two questions. First, does adaptive beat static (\ours{} vs.\ practitioner/random/grid search)? Second, does the fixed workflow with tree search outperform a general-purpose LLM agent (\ours{} vs.\ skill-based LLM agent)? \ours{} outperforms all static baselines on every task. Notably, the WildSci practitioner config scores below the base model (53.2\% vs.\ 53.6\%), yet \ours{} recovers to 58.5\%. \revised{Capping \ours{} at a matched 8 iterations (within the skill agent's 6--9 range), it still matches or exceeds the skill agent on all four tasks (Appendix~\ref{app:matched_budget}).} \revised{Appendix~\ref{app:main_analysis} collects further analysis of these results, including the detailed 8B OOM recovery, a full proposer reasoning trace, and failure modes beyond compute limits.}

\begin{figure}[t]
    \centering
    \includegraphics[width=\columnwidth]{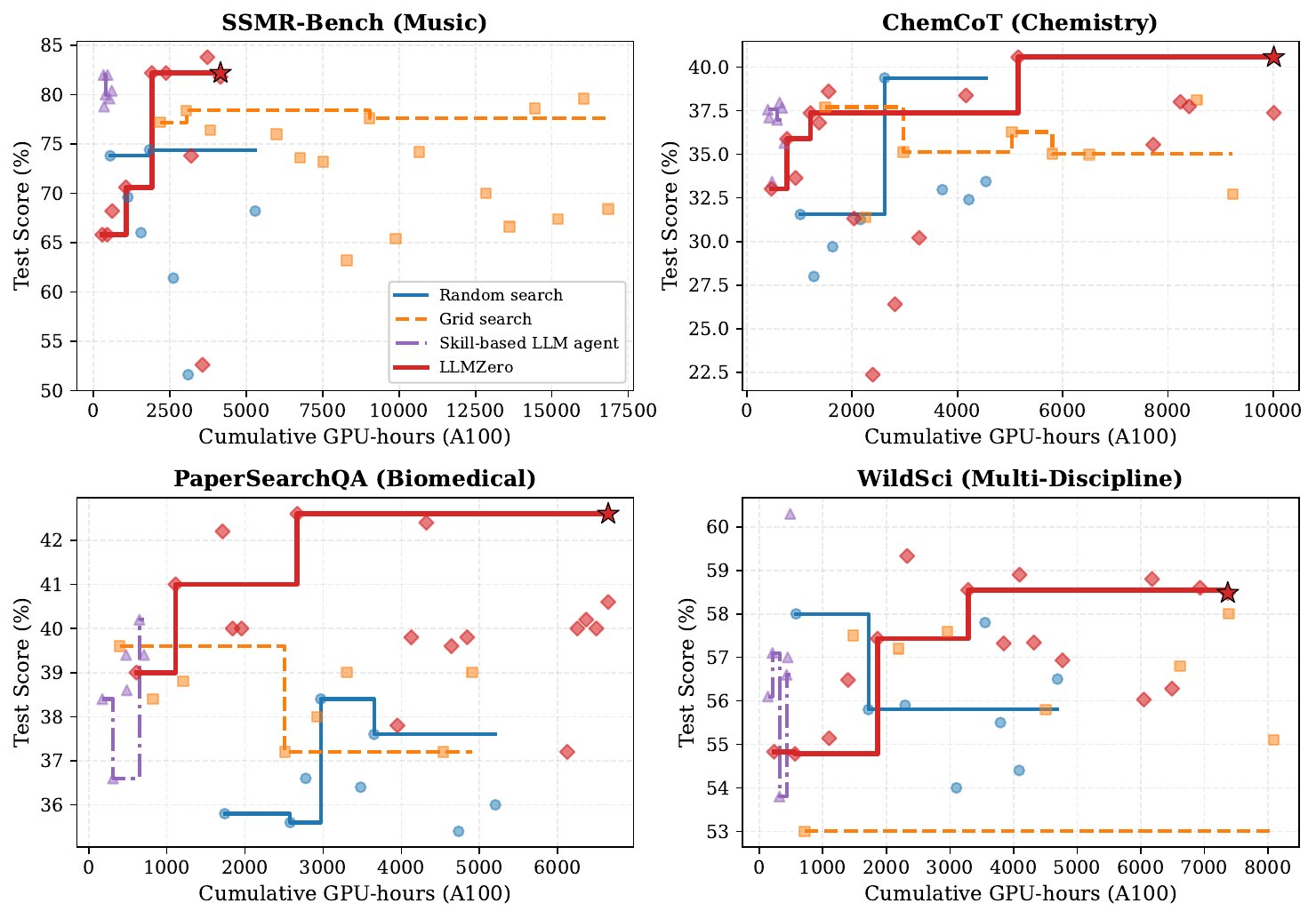}
    \caption{Test score at the best-validation run so far vs.\ cumulative GPU-hours. Dots show per-run/node test scores; step curves track the test score of whichever run has the highest validation so far.}
    \label{fig:compute}
\end{figure}

\paragraph{Compute efficiency.} Figure~\ref{fig:compute} plots best-so-far test score against cumulative GPU-hours for each method. \ours{} reaches the highest final test score on all 4 tasks under comparable total GPU compute (4{,}159--10{,}013 GPU-hours for \ours{} vs.\ 4{,}543--16{,}846 for HPO baselines). Early stopping terminates 56--70\% of nodes before completion. The forced from-scratch injection (\S\ref{sec:checkpoint}) explains why \ours{} uses more total training time than the skill-based agent.

\paragraph{API cost.}\label{sec:api_cost} \ours{}'s fixed workflow consumes \revised{73$\times$ less API cost overall (44--144$\times$ per task)} than the skill-based agent (\$48 vs.\ \$3{,}545 total; Table~\ref{tab:api_cost} in Appendix~\ref{app:api_cost}).

\begin{figure*}[t]
    \centering
    \begin{subfigure}[t]{0.49\textwidth}
        \centering
        \includegraphics[width=\textwidth]{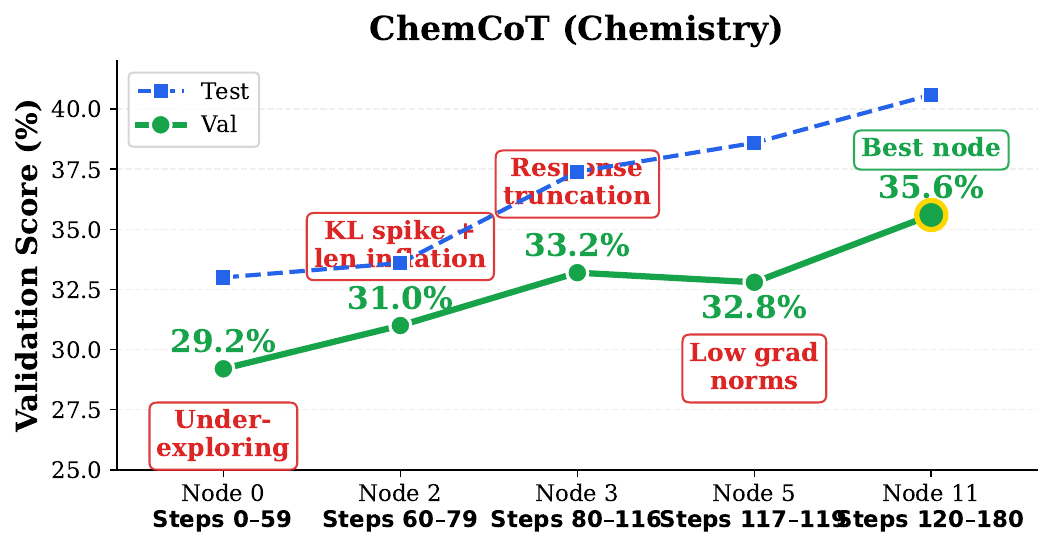}
    \end{subfigure}
    \hfill
    \begin{subfigure}[t]{0.49\textwidth}
        \centering
        \includegraphics[width=\textwidth]{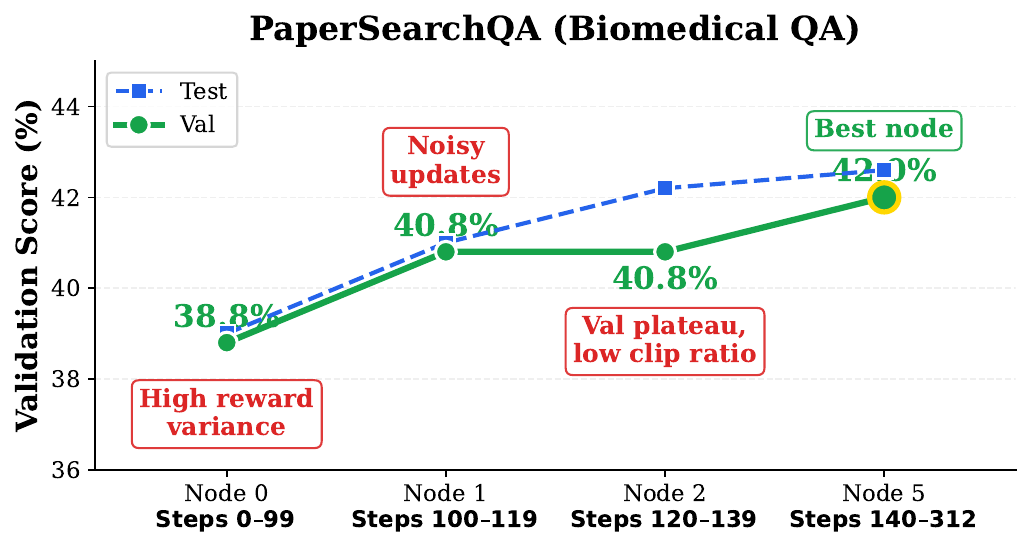}
    \end{subfigure}\\[-0.5em]
    \begin{subfigure}[t]{0.49\textwidth}
        \centering
        \includegraphics[width=\textwidth]{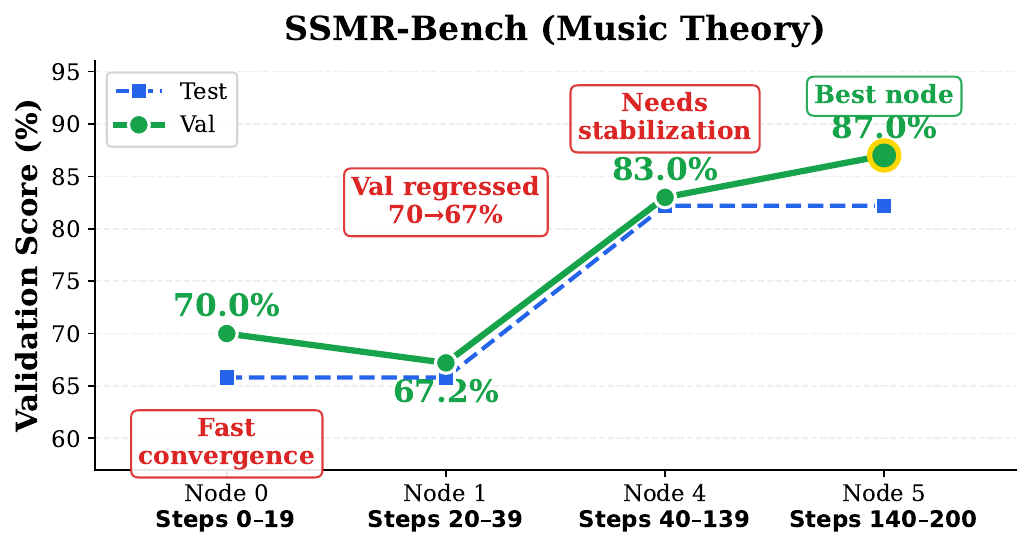}
    \end{subfigure}
    \hfill
    \begin{subfigure}[t]{0.49\textwidth}
        \centering
        \includegraphics[width=\textwidth]{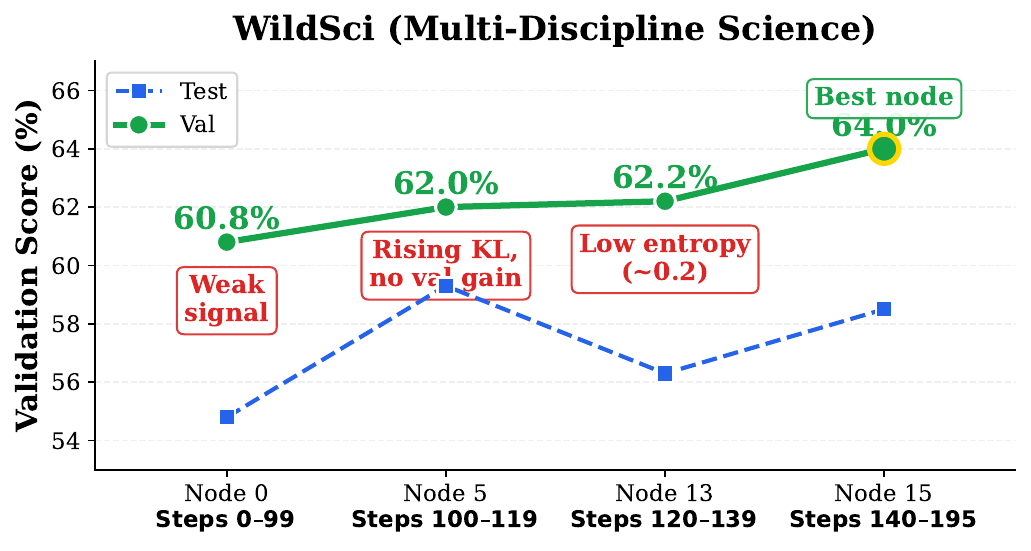}
    \end{subfigure}
    \caption{Best adaptive strategies across all four tasks. Green solid: validation score. Blue dashed: test score. Each point is one phase with annotations summarizing observed training dynamics.}
    \label{fig:all_strategies}
\end{figure*}

\subsection{Analysis of Discovered Strategies (RQ2)}
\label{sec:strategy_analysis}

We examine whether the adaptive strategies \ours{} discovers are dataset-dependent and what structural patterns emerge. Figure~\ref{fig:all_strategies} visualizes the best strategy for each task. We analyze ChemCoT and SSMR-Bench below; the PaperSearchQA and WildSci strategies are described in Appendix~\ref{app:extra_strategies}.

\subsubsection{Per-Dataset Strategies}
\label{sec:per_dataset}

\paragraph{ChemCoT (Chemistry).} The best strategy is a 5-phase chain (Figure~\ref{fig:all_strategies}). After initial training (val=29.2\%), the system widens the clip range and enables advantage normalization while lowering temperature (Phase~2, val=31.0\%), then observes a 16$\times$ KL loss spike coinciding with response length inflation (1630$\to$2278 tokens) without validation improvement, responding with a 5$\times$ KL penalty increase (Phase~3, val=33.2\%). This KL divergence spike preceded validation degradation by 1--2 phases, suggesting KL as a leading indicator that practitioners should monitor proactively. Phase~4 expands response capacity (6144$\to$7168 tokens) with a transient regression (val=32.8\%), before Phase~5 addresses low gradient norms ($<$0.001) and flat validation by raising LR and rollouts while reducing batch size (val=35.6\%).

\paragraph{Failure case: molecular optimization.} Nine subtasks (all \texttt{mol\_opt\_*}, \texttt{rxn\_retro}, \texttt{rxn\_nepp}, \texttt{rxn\_mechanism}) score 0\% across all methods including \ours{}. Molecular optimization requires valid SMILES string generation, which is a learned structural capability rather than a reasoning capability. No training strategy can elicit a skill the base model fundamentally lacks at 4B scale. This demonstrates the boundary of adaptive scheduling: it optimizes \emph{how} to train but cannot compensate for missing model capacity. \revised{See Appendix~\ref{app:failure_modes} for more details and further failure analysis.}

\paragraph{SSMR-Bench (Music Theory).} The best strategy is a 4-phase chain (Figure~\ref{fig:all_strategies}). From Phase~1 (val=70.0\%), Phase~2 applies multiple conservative changes simultaneously: reducing LR to 3e-5, doubling KL to 0.002, reducing gradient clipping to 0.5, enabling advantage normalization, and raising temperature to 1.1. This over-constrains learning and causes regression to 67.2\%. Phase~3 reverses the core constraints (restoring LR to 5e-5, relaxing KL to 0.001, widening clip to 0.30) while retaining the higher temperature, driving a +15.8~pp recovery (val=83.0\%). Phase~4 re-tightens for convergence (LR=3e-5, KL=0.002, T=1.0) with expanded response length (6144$\to$7168 tokens), reaching val=87.0\% (test=82.2\%). LR and KL oscillate across phases while epochs and response capacity accumulate monotonically. The agentic early stopper terminated 7 of 10 explored nodes.

\subsubsection{Cross-Task Structural Patterns}
\label{sec:patterns}

Three empirical observations emerge from comparing strategies across tasks:

\paragraph{Dataset-specific dynamics determine strategy structure.} Each task shows a distinct observable pattern: KL divergence spikes with response-length inflation (ChemCoT), stagnating validation with near-zero clip ratios (PaperSearchQA), validation regression under conservative hyperparameters (SSMR-Bench), and model collapse after aggressive LR scaling (WildSci). These emerge unpredictably (ChemCoT's KL spike appears only after 2 phases) and require severity-calibrated interventions, which the KL trajectories mirror: reactive spikes, monotonic increase, symmetric oscillation, and tighten-then-relax, respectively. No fixed schedule can anticipate which pattern will dominate or when.

\paragraph{Multi-dimensional transitions are effective.} In all 4 tasks, the highest-gain transition changes 3+ hyperparameters in coordinated combinations; e.g., ChemCoT Phase~5 jointly raises LR and KL penalty, which the proposer reasoned would escape the plateau without diverging. Such interventions are unlikely to be found by tuning parameters independently.

Notably, the KL coefficient is the most frequently adjusted parameter (changed in 12 of 13 transitions), yet it is held constant in all surveyed multi-stage works (\S\ref{sec:intro}). Its non-monotonic trajectory appears load-bearing: the best strategies tighten KL reactively and relax it proactively, so a fixed KL schedule would miss these dynamics on most tasks.

\paragraph{Capacity parameters accumulate while regularization parameters oscillate.} Across all 4 best strategies, response length and rollout count show \emph{zero} direction reversals, while LR and temperature reverse 1--2 times on every task; the KL coefficient reverses on 3 of 4 tasks but rises monotonically on PaperSearchQA, where it progressively tightens to stabilize noisy updates. We read this as a candidate design heuristic: \emph{capacity parameters should accumulate monotonically while regularization parameters should be free to oscillate in response to shifting training dynamics.} The asymmetry \revised{may reflect} that capacity parameters are information-constructive (reducing them truncates reasoning chains or raises gradient variance), while regularization parameters control an exploration-exploitation tradeoff whose optimal balance shifts continuously. \revised{The pattern is not an artifact of directional prompt cues: of all metric-tracked hyperparameters (Appendix~\ref{app:hp_desc}), only the KL coefficient and the maximum response length carry a single directional hint, yet behavior splits by parameter class, not hint direction. KL oscillates despite a hint suggesting only conditional increase, and monotonic response-length accumulation is independently corroborated by prior multi-stage RL work~\citep{deepscaler,acereason,deepcoder,mimo}. An off-best-path ChemCoT node did reduce capacity parameters (rollouts 8$\to$4), so the search space is not one-directional. We nonetheless scope the heuristic to our evaluated setting; its generality across architectures, optimizers, reward structures, dataset sizes, and RL algorithms remains to be established.}

\paragraph{Diagnosis quality.} We human-verified every observation in the proposer's diagnoses against metric traces (Tables~\ref{tab:strategy_chemcot}--\ref{tab:strategy_wildsci}); all are correct. Of 13 non-initial transitions, 11 improve validation. On 3 of 4 tasks test scores increase monotonically along the best path (Figure~\ref{fig:all_strategies}); on WildSci one transition regresses but the next recovers.

\subsection{Scaling Analysis (RQ3)}
\label{sec:scaling}

\begin{figure}[t]
\centering
\begin{tikzpicture}
\begin{axis}[
    width=\columnwidth,
    height=5cm,
    ybar,
    bar width=6pt,
    ylabel={Average Accuracy (\%)},
    symbolic x coords={0.6B, 1.7B, 4B, 8B},
    xtick=data,
    xlabel={Model Size},
    ymin=0, ymax=100,
    legend style={at={(0.5,1.05)}, anchor=south, font=\small, cells={anchor=west}},
    legend columns=3,
    nodes near coords style={font=\tiny, rotate=90, anchor=west},
    every node near coord/.append style={yshift=1pt},
    enlarge x limits=0.2,
    grid=major,
    grid style={dashed, gray!30},
]
\addplot[fill=gray!40, nodes near coords] coordinates {(0.6B,32.4) (1.7B,31.6) (4B,51.4) (8B,52.6)};
\addplot[fill=blue!40, nodes near coords={\pgfplotspointmeta}, point meta=explicit symbolic] coordinates {(0.6B,53.0) [53.0] (1.7B,52.2) [52.2] (4B,65.8) [65.8] (8B,0) [OOM]};
\addplot[fill=red!60, nodes near coords] coordinates {(0.6B,72.4) (1.7B,62.4) (4B,82.2) (8B,83.6)};
\legend{Qwen3 (base), Practitioner, \ours{}}
\end{axis}
\end{tikzpicture}
\caption{Model scaling on SSMR-Bench (average across 4 subtasks). Practitioner config failed (OOM) on 8B. Per-subtask breakdown in Table~\ref{tab:scaling} (Appendix~\ref{app:scaling}).}
\label{fig:scaling}
\end{figure}
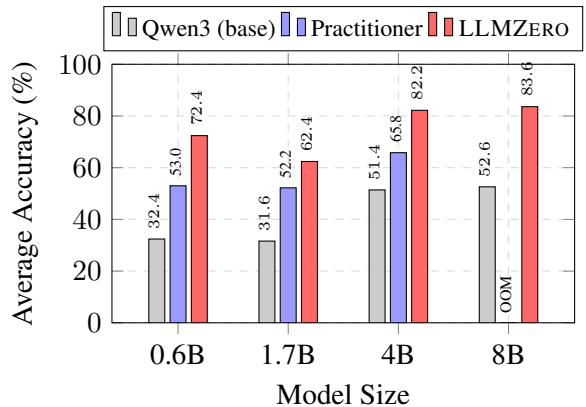

Figure~\ref{fig:scaling} reports average accuracy on SSMR-Bench across model sizes. \ours{} consistently outperforms baselines from 0.6B to 8B, with gains of +30.8~pp to +40.0~pp over the base model. The practitioner config failed with OOM on 8B, while \ours{} autonomously discovered a working configuration (83.6\%). \revised{Note that OOM recovery is not a fundamental capability of the method, since a practitioner could resolve it manually by adjusting memory settings. \ours{} navigated the failure autonomously as a proof of its robustness. A detailed breakdown of the diagnosis and the decisive configuration changes is in Appendix~\ref{app:8b_oom}.}

\subsection{Ablation Studies (RQ4)}
\label{sec:ablation}

Table~\ref{tab:ablation} ablates key components on SSMR-Bench to distinguish which drive accuracy gains versus compute efficiency. Removing multi-stage composition drops accuracy by 9.4~pp (82.2\%$\to$72.8\%), confirming that the ability to compose adaptive multi-stage strategies is the primary driver of improvement; without it, the system reduces to selecting the best single-phase configuration from the same search budget. Removing visual reasoning or early stopping yields on-par accuracy, but at substantially worse compute efficiency. Visual reasoning enables the early stopper to reliably judge trajectory dominance from overlay plots, while early stopping itself focuses compute on promising branches. Together they reduce wall-clock time without sacrificing strategy quality. \revised{The small accuracy gaps from removing visual reasoning or early stopping sit within single-seed variance, and both ablations leave the LLM's text-based reading of dynamics active, so neither isolates the reasoning itself. A reward-only ablation that strips all diagnostic signals does: it drops the test mean from 61.5\% to 49.5\% on the public VeRL 0.7.1 stack (Appendix~\ref{app:reward_only}); removing the proposer entirely would collapse \ours{} into the random/grid baselines it already outperforms.} \revised{See Appendix~\ref{app:public_stack} for further ablation studies on the public VeRL 0.7.1 stack, including backbone-quality and poor-default from-scratch analyses.}

\subsection{Strategy Transfer (RQ5)}
\label{sec:knowledge}

We test whether multi-stage structure alone drives improvement by executing two fixed schedules on three held-out tasks (Table~\ref{tab:transfer}): the discovered SSMR-Bench 4-phase strategy, and a Capacity Guidebook that progressively increases only response length and rollout count following the community practice~\citep{deepscaler,acereason} (Table~\ref{tab:guidebook_config}), with stage durations estimated from our successful trajectories. Both fixed schedules improve over the base model on all tasks (+4.6 to +24.4~pp for SSMR transfer, +4.8 to +19.7~pp for the Guidebook), confirming that multi-stage training is broadly beneficial. However, their gains are inconsistent: the SSMR transfer nearly matches adaptive search on WildSci (58.3\% vs.\ 58.5\%) but underperforms by 6.4~pp on PaperSearchQA, while the Guidebook lags the SSMR transfer by 4.7~pp on ChemCoT. These inconsistencies \revised{suggest that} fixed schedules \revised{do not always reliably generalize} across tasks without adapting to observed training dynamics. Notably, the strong transfer performance on some tasks may benefit from identical base model and dataset sizes, which produce similar training dynamics; adaptive search remains necessary for robustness when dataset scale or model family varies (More analysis in Appendix~\ref{app:guidebook_analysis}).

\begin{table}[t]
\caption{Ablation on SSMR-Bench: test accuracy (\%) at best validation node. Speed is relative compute efficiency normalized to the full system.}
\label{tab:ablation}
\centering
\scalebox{0.84}{%
\small
\begin{tabular}{l*{4}{c}cc}
\toprule
\textbf{Variant} & {\scriptsize Scl} & {\scriptsize Bea} & {\scriptsize Cho} & {\scriptsize Int} & {\scriptsize Avg} & {\scriptsize Speed} \\
\midrule
\ours{} (full) & 94.4 & 81.6 & 77.6 & 75.2 & \textbf{82.2} & \textbf{1.0$\times$} \\
\midrule
w/o visual reasoning & 93.6 & 85.6 & 80.0 & 70.4 & 82.4 & 0.6$\times$ \\
w/o multi-stage & 88.8 & 78.4 & 62.4 & 61.6 & 72.8 & 0.22$\times$ \\
w/o early stopping & 93.6 & 90.4 & 75.2 & 72.0 & 82.8 & 0.29$\times$ \\
\bottomrule
\end{tabular}%
}
\end{table}

\begin{table}[t]
\caption{Strategy transfer: test scores (\%) for fixed multi-stage schedules on held-out tasks. SSMR transfer applies the discovered 4-phase strategy. Capacity Guidebook applies only progressive capacity scaling.}
\label{tab:transfer}
\centering
\resizebox{\columnwidth}{!}{%
\small
\begin{tabular}{lccccc}
\toprule
\textbf{Dataset} & \textbf{Practitioner} & \textbf{Random} & \textbf{SSMR transfer} & \textbf{Cap.\ Guidebook} & \textbf{\ours{}} \\
\midrule
PaperSearchQA & 39.0 & 37.6 & 36.2 & 37.8 & \textbf{42.6} \\
WildSci       & 53.2 & 55.8 & \textbf{58.3} & \textbf{58.4} & \textbf{58.5} \\
ChemCoT       & 33.0 & 39.4 & \textbf{41.3} & 36.6 & 40.5 \\
\midrule
Average       & 41.7 & 44.3 & 45.3 & 44.3 & \textbf{47.2} \\
\bottomrule
\end{tabular}%
}
\end{table}

\section{Conclusion}
\label{sec:conclusion}


\revised{We observe a recurring structural asymmetry in multi-stage GRPO strategies}: capacity parameters (e.g., response length, rollouts) accumulate monotonically, while regularization parameters (e.g., learning rate, KL coefficient, temperature) predominantly oscillate in response to shifting training dynamics. We uncover this through \ours{}, an agentic system that reasons about training dynamics at each checkpoint and proposes coordinated, multi-parameter transitions to address diagnosed pathologies. Across four diverse tasks, these adaptive strategies improve over the base model by 9\% to 140\% and over grid search by 6\% to 15\% (relative), outperforming all baselines. These findings suggest that the current paradigm's narrow focus on staging one or two capacity parameters may constrain models' potential, and that dynamics-aware, multi-dimensional adaptation is beneficial in RL post-training. We leave the extension of this approach beyond GRPO, including to other RL finetuning algorithms, supervised finetuning, and foundation model pretraining, to future work.

\section*{Limitations}

Our system expands one node at a time; a hybrid with population-based training~\citep{pbt} that maintains multiple trajectories with LLM-guided transitions would combine broad exploration with intelligent proposals and is a natural next step. All datasets are subsampled to 5{,}000 training examples to keep per-iteration search tractable; validating that discovered strategies and structural patterns hold at production data scales and model scales requires substantially more compute and is left to future work.


\bibliography{references}

@misc{dpo,
      title={Direct Preference Optimization: Your Language Model is Secretly a Reward Model}, 
      author={Rafael Rafailov and Archit Sharma and Eric Mitchell and Stefano Ermon and Christopher D. Manning and Chelsea Finn},
      year={2024},
      eprint={2305.18290},
      archivePrefix={arXiv},
      primaryClass={cs.LG},
      url={https://arxiv.org/abs/2305.18290}, 
}

@misc{dsmath,
      title={DeepSeekMath: Pushing the Limits of Mathematical Reasoning in Open Language Models}, 
      author={Zhihong Shao and Peiyi Wang and Qihao Zhu and Runxin Xu and Junxiao Song and Xiao Bi and Haowei Zhang and Mingchuan Zhang and Y. K. Li and Y. Wu and Daya Guo},
      year={2024},
      eprint={2402.03300},
      archivePrefix={arXiv},
      primaryClass={cs.CL},
      url={https://arxiv.org/abs/2402.03300}, 
}

@article{dsr1,
   title={DeepSeek-R1 incentivizes reasoning in LLMs through reinforcement learning},
   volume={645},
   ISSN={1476-4687},
   url={http://dx.doi.org/10.1038/s41586-025-09422-z},
   DOI={10.1038/s41586-025-09422-z},
   number={8081},
   journal={Nature},
   publisher={Springer Science and Business Media LLC},
   author={Guo, Daya and Yang, Dejian and Zhang, Haowei and Song, Junxiao and Wang, Peiyi and Zhu, Qihao and Xu, Runxin and Zhang, Ruoyu and Ma, Shirong and Bi, Xiao and Zhang, Xiaokang and Yu, Xingkai and Wu, Yu and Wu, Z. F. and Gou, Zhibin and Shao, Zhihong and Li, Zhuoshu and Gao, Ziyi and Liu, Aixin and Xue, Bing and Wang, Bingxuan and Wu, Bochao and Feng, Bei and Lu, Chengda and Zhao, Chenggang and Deng, Chengqi and Ruan, Chong and Dai, Damai and Chen, Deli and Ji, Dongjie and Li, Erhang and Lin, Fangyun and Dai, Fucong and Luo, Fuli and Hao, Guangbo and Chen, Guanting and Li, Guowei and Zhang, H. and Xu, Hanwei and Ding, Honghui and Gao, Huazuo and Qu, Hui and Li, Hui and Guo, Jianzhong and Li, Jiashi and Chen, Jingchang and Yuan, Jingyang and Tu, Jinhao and Qiu, Junjie and Li, Junlong and Cai, J. L. and Ni, Jiaqi and Liang, Jian and Chen, Jin and Dong, Kai and Hu, Kai and You, Kaichao and Gao, Kaige and Guan, Kang and Huang, Kexin and Yu, Kuai and Wang, Lean and Zhang, Lecong and Zhao, Liang and Wang, Litong and Zhang, Liyue and Xu, Lei and Xia, Leyi and Zhang, Mingchuan and Zhang, Minghua and Tang, Minghui and Zhou, Mingxu and Li, Meng and Wang, Miaojun and Li, Mingming and Tian, Ning and Huang, Panpan and Zhang, Peng and Wang, Qiancheng and Chen, Qinyu and Du, Qiushi and Ge, Ruiqi and Zhang, Ruisong and Pan, Ruizhe and Wang, Runji and Chen, R. J. and Jin, R. L. and Chen, Ruyi and Lu, Shanghao and Zhou, Shangyan and Chen, Shanhuang and Ye, Shengfeng and Wang, Shiyu and Yu, Shuiping and Zhou, Shunfeng and Pan, Shuting and Li, S. S. and Zhou, Shuang and Wu, Shaoqing and Yun, Tao and Pei, Tian and Sun, Tianyu and Wang, T. and Zeng, Wangding and Liu, Wen and Liang, Wenfeng and Gao, Wenjun and Yu, Wenqin and Zhang, Wentao and Xiao, W. L. and An, Wei and Liu, Xiaodong and Wang, Xiaohan and Chen, Xiaokang and Nie, Xiaotao and Cheng, Xin and Liu, Xin and Xie, Xin and Liu, Xingchao and Yang, Xinyu and Li, Xinyuan and Su, Xuecheng and Lin, Xuheng and Li, X. Q. and Jin, Xiangyue and Shen, Xiaojin and Chen, Xiaosha and Sun, Xiaowen and Wang, Xiaoxiang and Song, Xinnan and Zhou, Xinyi and Wang, Xianzu and Shan, Xinxia and Li, Y. K. and Wang, Y. Q. and Wei, Y. X. and Zhang, Yang and Xu, Yanhong and Li, Yao and Zhao, Yao and Sun, Yaofeng and Wang, Yaohui and Yu, Yi and Zhang, Yichao and Shi, Yifan and Xiong, Yiliang and He, Ying and Piao, Yishi and Wang, Yisong and Tan, Yixuan and Ma, Yiyang and Liu, Yiyuan and Guo, Yongqiang and Ou, Yuan and Wang, Yuduan and Gong, Yue and Zou, Yuheng and He, Yujia and Xiong, Yunfan and Luo, Yuxiang and You, Yuxiang and Liu, Yuxuan and Zhou, Yuyang and Zhu, Y. X. and Huang, Yanping and Li, Yaohui and Zheng, Yi and Zhu, Yuchen and Ma, Yunxian and Tang, Ying and Zha, Yukun and Yan, Yuting and Ren, Z. Z. and Ren, Zehui and Sha, Zhangli and Fu, Zhe and Xu, Zhean and Xie, Zhenda and Zhang, Zhengyan and Hao, Zhewen and Ma, Zhicheng and Yan, Zhigang and Wu, Zhiyu and Gu, Zihui and Zhu, Zijia and Liu, Zijun and Li, Zilin and Xie, Ziwei and Song, Ziyang and Pan, Zizheng and Huang, Zhen and Xu, Zhipeng and Zhang, Zhongyu and Zhang, Zhen},
   year={2025},
   month=Sept, pages={633–638} }

@article{randomhpo,
  author  = {James Bergstra and Yoshua Bengio},
  title   = {Random Search for Hyper-Parameter Optimization},
  journal = {Journal of Machine Learning Research},
  year    = {2012},
  volume  = {13},
  number  = {10},
  pages   = {281--305},
  url     = {http://jmlr.org/papers/v13/bergstra12a.html}
}

@misc{bayesianhpo,
      title={Practical Bayesian Optimization of Machine Learning Algorithms}, 
      author={Jasper Snoek and Hugo Larochelle and Ryan P. Adams},
      year={2012},
      eprint={1206.2944},
      archivePrefix={arXiv},
      primaryClass={stat.ML},
      url={https://arxiv.org/abs/1206.2944}, 
}

@misc{hyperpod,
      title={Hyperband: A Novel Bandit-Based Approach to Hyperparameter Optimization}, 
      author={Lisha Li and Kevin Jamieson and Giulia DeSalvo and Afshin Rostamizadeh and Ameet Talwalkar},
      year={2018},
      eprint={1603.06560},
      archivePrefix={arXiv},
      primaryClass={cs.LG},
      url={https://arxiv.org/abs/1603.06560}, 
}

@misc{pbt,
      title={Population Based Training of Neural Networks}, 
      author={Max Jaderberg and Valentin Dalibard and Simon Osindero and Wojciech M. Czarnecki and Jeff Donahue and Ali Razavi and Oriol Vinyals and Tim Green and Iain Dunning and Karen Simonyan and Chrisantha Fernando and Koray Kavukcuoglu},
      year={2017},
      eprint={1711.09846},
      archivePrefix={arXiv},
      primaryClass={cs.LG},
      url={https://arxiv.org/abs/1711.09846}, 
}

@misc{aide,
      title={AIDE: AI-Driven Exploration in the Space of Code}, 
      author={Zhengyao Jiang and Dominik Schmidt and Dhruv Srikanth and Dixing Xu and Ian Kaplan and Deniss Jacenko and Yuxiang Wu},
      year={2025},
      eprint={2502.13138},
      archivePrefix={arXiv},
      primaryClass={cs.AI},
      url={https://arxiv.org/abs/2502.13138}, 
}

@misc{sela,
      title={SELA: Tree-Search Enhanced LLM Agents for Automated Machine Learning}, 
      author={Yizhou Chi and Yizhang Lin and Sirui Hong and Duyi Pan and Yaying Fei and Guanghao Mei and Bangbang Liu and Tianqi Pang and Jacky Kwok and Ceyao Zhang and Bang Liu and Chenglin Wu},
      year={2024},
      eprint={2410.17238},
      archivePrefix={arXiv},
      primaryClass={cs.AI},
      url={https://arxiv.org/abs/2410.17238}, 
}

@misc{alphaevolve,
      title={AlphaEvolve: A coding agent for scientific and algorithmic discovery}, 
      author={Alexander Novikov and Ngân Vũ and Marvin Eisenberger and Emilien Dupont and Po-Sen Huang and Adam Zsolt Wagner and Sergey Shirobokov and Borislav Kozlovskii and Francisco J. R. Ruiz and Abbas Mehrabian and M. Pawan Kumar and Abigail See and Swarat Chaudhuri and George Holland and Alex Davies and Sebastian Nowozin and Pushmeet Kohli and Matej Balog},
      year={2025},
      eprint={2506.13131},
      archivePrefix={arXiv},
      primaryClass={cs.AI},
      url={https://arxiv.org/abs/2506.13131}, 
}

@misc{tot,
      title={Tree of Thoughts: Deliberate Problem Solving with Large Language Models}, 
      author={Shunyu Yao and Dian Yu and Jeffrey Zhao and Izhak Shafran and Thomas L. Griffiths and Yuan Cao and Karthik Narasimhan},
      year={2023},
      eprint={2305.10601},
      archivePrefix={arXiv},
      primaryClass={cs.CL},
      url={https://arxiv.org/abs/2305.10601}, 
}

@misc{hao2023reasoning,
      title={Reasoning with Language Model is Planning with World Model}, 
      author={Shibo Hao and Yi Gu and Haodi Ma and Joshua Jiahua Hong and Zhen Wang and Daisy Zhe Wang and Zhiting Hu},
      year={2023},
      eprint={2305.14992},
      archivePrefix={arXiv},
      primaryClass={cs.CL},
      url={https://arxiv.org/abs/2305.14992}, 
}

@inproceedings{uct,
author = {Kocsis, Levente and Szepesv\'{a}ri, Csaba},
title = {Bandit based monte-carlo planning},
year = {2006},
isbn = {354045375X},
publisher = {Springer-Verlag},
address = {Berlin, Heidelberg},
url = {https://doi.org/10.1007/11871842_29},
doi = {10.1007/11871842_29},
booktitle = {Proceedings of the 17th European Conference on Machine Learning},
pages = {282–293},
numpages = {12},
location = {Berlin, Germany},
series = {ECML'06}
}

@misc{mlzero,
      title={MLZero: A Multi-Agent System for End-to-end Machine Learning Automation}, 
      author={Haoyang Fang and Boran Han and Nick Erickson and Xiyuan Zhang and Su Zhou and Anirudh Dagar and Jiani Zhang and Ali Caner Turkmen and Cuixiong Hu and Huzefa Rangwala and Ying Nian Wu and Bernie Wang and George Karypis},
      year={2025},
      eprint={2505.13941},
      archivePrefix={arXiv},
      primaryClass={cs.MA},
      url={https://arxiv.org/abs/2505.13941}, 
}

@misc{deepscaler,
  title={DeepScaleR: Surpassing O1-Preview with a 1.5B Model by Scaling RL},
  author={Michael Luo and Sijun Tan and Justin Wong and Xiaoxiang Shi and William Y. Tang and Manan Roongta and Colin Cai and Jeffrey Luo and Li Erran Li and Raluca Ada Popa and Ion Stoica},
  note={Notion Blog},
  year={2025}
}

@misc{acereason,
      title={AceReason-Nemotron: Advancing Math and Code Reasoning through Reinforcement Learning}, 
      author={Yang Chen and Zhuolin Yang and Zihan Liu and Chankyu Lee and Peng Xu and Mohammad Shoeybi and Bryan Catanzaro and Wei Ping},
      year={2025},
      eprint={2505.16400},
      archivePrefix={arXiv},
      primaryClass={cs.LG},
      url={https://arxiv.org/abs/2505.16400}, 
}

@misc{fastcurl,
      title={FastCuRL: Curriculum Reinforcement Learning with Stage-wise Context Scaling for Efficient Training R1-like Reasoning Models}, 
      author={Mingyang Song and Mao Zheng and Zheng Li and Wenjie Yang and Xuan Luo and Yue Pan and Feng Zhang},
      year={2025},
      eprint={2503.17287},
      archivePrefix={arXiv},
      primaryClass={cs.CL},
      url={https://arxiv.org/abs/2503.17287}, 
}

@misc{skywork_or1,
      title={Skywork Open Reasoner 1 Technical Report}, 
      author={Jujie He and Jiacai Liu and Chris Yuhao Liu and Rui Yan and Chaojie Wang and Peng Cheng and Xiaoyu Zhang and Fuxiang Zhang and Jiacheng Xu and Wei Shen and Siyuan Li and Liang Zeng and Tianwen Wei and Cheng Cheng and Bo An and Yang Liu and Yahui Zhou},
      year={2025},
      eprint={2505.22312},
      archivePrefix={arXiv},
      primaryClass={cs.LG},
      url={https://arxiv.org/abs/2505.22312}, 
}

@misc{jt_math,
      title={JT-Math: A Multi-Stage Framework for Advanced Mathematical Reasoning in Large Language Models}, 
      author={Yifan Hao and Fangning Chao and Yaqian Hao and Zhaojun Cui and Huan Bai and Haiyu Zhang and Yankai Liu and Chao Deng and Junlan Feng},
      year={2025},
      eprint={2507.19748},
      archivePrefix={arXiv},
      primaryClass={cs.CL},
      url={https://arxiv.org/abs/2507.19748}, 
}

@misc{mimo,
      title={MiMo: Unlocking the Reasoning Potential of Language Model -- From Pretraining to Posttraining}, 
      author={LLM-Core Xiaomi and : and Bingquan Xia and Bowen Shen and Cici and Dawei Zhu and Di Zhang and Gang Wang and Hailin Zhang and Huaqiu Liu and Jiebao Xiao and Jinhao Dong and Liang Zhao and Peidian Li and Peng Wang and Shihua Yu and Shimao Chen and Weikun Wang and Wenhan Ma and Xiangwei Deng and Yi Huang and Yifan Song and Zihan Jiang and Bowen Ye and Can Cai and Chenhong He and Dong Zhang and Duo Zhang and Guoan Wang and Hao Tian and Haochen Zhao and Heng Qu and Hongshen Xu and Jun Shi and Kainan Bao and Kai Fang and Kang Zhou and Kangyang Zhou and Lei Li and Menghang Zhu and Nuo Chen and Qiantong Wang and Shaohui Liu and Shicheng Li and Shuhao Gu and Shuhuai Ren and Shuo Liu and Sirui Deng and Weiji Zhuang and Weiwei Lv and Wenyu Yang and Xin Zhang and Xing Yong and Xing Zhang and Xingchen Song and Xinzhe Xu and Xu Wang and Yihan Yan and Yu Tu and Yuanyuan Tian and Yudong Wang and Yue Yu and Zhenru Lin and Zhichao Song and Zihao Yue},
      year={2025},
      eprint={2505.07608},
      archivePrefix={arXiv},
      primaryClass={cs.CL},
      url={https://arxiv.org/abs/2505.07608}, 
}

@misc{deepcoder,
  title={DeepCoder: A Fully Open-Source 14B Coder at O3-mini Level},
  author={Michael Luo and Sijun Tan and Roy Huang and Ameen Patel and Alpay Ariyak and Qingyang Wu and Xiaoxiang Shi and Rachel Xin and Colin Cai and Maurice Weber and Ce Zhang and Li Erran Li and Raluca Ada Popa and Ion Stoica},
  note={Notion Blog},
  year={2025}
}

@misc{still3,
      title={An Empirical Study on Eliciting and Improving R1-like Reasoning Models}, 
      author={Zhipeng Chen and Yingqian Min and Beichen Zhang and Jie Chen and Jinhao Jiang and Daixuan Cheng and Wayne Xin Zhao and Zheng Liu and Xu Miao and Yang Lu and Lei Fang and Zhongyuan Wang and Ji-Rong Wen},
      year={2025},
      eprint={2503.04548},
      archivePrefix={arXiv},
      primaryClass={cs.CL},
      url={https://arxiv.org/abs/2503.04548}, 
}

@misc{tacler,
      title={TACLer: Tailored Curriculum Reinforcement Learning for Efficient Reasoning}, 
      author={Huiyuan Lai and Malvina Nissim},
      year={2026},
      eprint={2601.21711},
      archivePrefix={arXiv},
      primaryClass={cs.CL},
      url={https://arxiv.org/abs/2601.21711}, 
}

@misc{siri,
      title={SIRI: Scaling Iterative Reinforcement Learning with Interleaved Compression}, 
      author={Haoming Wen and Yushi Bai and Juanzi Li and Jie Tang},
      year={2025},
      eprint={2509.25176},
      archivePrefix={arXiv},
      primaryClass={cs.LG},
      url={https://arxiv.org/abs/2509.25176}, 
}

@misc{qwenlong_l1,
      title={QwenLong-L1: Towards Long-Context Large Reasoning Models with Reinforcement Learning}, 
      author={Fanqi Wan and Weizhou Shen and Shengyi Liao and Yingcheng Shi and Chenliang Li and Ziyi Yang and Ji Zhang and Fei Huang and Jingren Zhou and Ming Yan},
      year={2025},
      eprint={2505.17667},
      archivePrefix={arXiv},
      primaryClass={cs.CL},
      url={https://arxiv.org/abs/2505.17667}, 
}

@misc{p1vl,
      title={P1-VL: Bridging Visual Perception and Scientific Reasoning in Physics Olympiads}, 
      author={Yun Luo and Futing Wang and Qianjia Cheng and Fangchen Yu and Haodi Lei and Jianhao Yan and Chenxi Li and Jiacheng Chen and Yufeng Zhao and Haiyuan Wan and Yuchen Zhang and Shenghe Zheng and Junchi Yao and Qingyang Zhang and Haonan He and Wenxuan Zeng and Li Sheng and Chengxing Xie and Yuxin Zuo and Yizhuo Li and Yulun Wu and Rui Huang and Dongzhan Zhou and Kai Chen and Yu Qiao and Lei Bai and Yu Cheng and Ning Ding and Bowen Zhou and Peng Ye and Ganqu Cui},
      year={2026},
      eprint={2602.09443},
      archivePrefix={arXiv},
      primaryClass={cs.AI},
      url={https://arxiv.org/abs/2602.09443}, 
}

@misc{diff_aware_staged_rl,
      title={How Difficulty-Aware Staged Reinforcement Learning Enhances LLMs' Reasoning Capabilities: A Preliminary Experimental Study}, 
      author={Yunjie Ji and Sitong Zhao and Xiaoyu Tian and Haotian Wang and Shuaiting Chen and Yiping Peng and Han Zhao and Xiangang Li},
      year={2025},
      eprint={2504.00829},
      archivePrefix={arXiv},
      primaryClass={cs.CL},
      url={https://arxiv.org/abs/2504.00829}, 
}

@misc{chemcotbench,
      title={Beyond Chemical QA: Evaluating LLM's Chemical Reasoning with Modular Chemical Operations}, 
      author={Hao Li and He Cao and Bin Feng and Yanjun Shao and Xiangru Tang and Zhiyuan Yan and Li Yuan and Yonghong Tian and Yu Li},
      year={2026},
      eprint={2505.21318},
      archivePrefix={arXiv},
      primaryClass={cs.AI},
      url={https://arxiv.org/abs/2505.21318}, 
}

@misc{papersearchqa,
      title={PaperSearchQA: Learning to Search and Reason over Scientific Papers with RLVR}, 
      author={James Burgess and Jan N. Hansen and Duo Peng and Yuhui Zhang and Alejandro Lozano and Min Woo Sun and Emma Lundberg and Serena Yeung-Levy},
      year={2026},
      eprint={2601.18207},
      archivePrefix={arXiv},
      primaryClass={cs.LG},
      url={https://arxiv.org/abs/2601.18207}, 
}

@misc{ssmrbench,
      title={Towards an AI Musician: Synthesizing Sheet Music Problems for Musical Reasoning}, 
      author={Zhilin Wang and Zhe Yang and Yun Luo and Yafu Li and Xiaoye Qu and Ziqian Qiao and Haoran Zhang and Runzhe Zhan and Derek F. Wong and Jizhe Zhou and Yu Cheng},
      year={2025},
      eprint={2509.04059},
      archivePrefix={arXiv},
      primaryClass={cs.CL},
      url={https://arxiv.org/abs/2509.04059}, 
}

@misc{wildsci,
      title={WildSci: Advancing Scientific Reasoning from In-the-Wild Literature}, 
      author={Tengxiao Liu and Deepak Nathani and Zekun Li and Kevin Yang and William Yang Wang},
      year={2026},
      eprint={2601.05567},
      archivePrefix={arXiv},
      primaryClass={cs.AI},
      url={https://arxiv.org/abs/2601.05567}, 
}

@article{posttrainbench,
  title={PostTrainBench: Can LLM Agents Automate LLM Post-Training?},
  author={Rank, Ben and Bhatnagar, Hardik and Prabhu, Ameya and Eisenberg, Shira and Nguyen, Karina and Bethge, Matthias and Andriushchenko, Maksym},
  journal={arXiv preprint arXiv:2603.08640},
  year={2026}
}

@article{datamaster,
  title={DataMaster: Towards Autonomous Data Engineering for Machine Learning},
  author={Du, Yaxin and Yang, Xiyuan and Zhou, Zhifan and Liu, Wanxu and Lei, Zixing and Chen, Zimeng and Liu, Fenyi and Wu, Haotian and Cai, Yuzhu and Liu, Zexi and others},
  journal={arXiv preprint arXiv:2605.10906},
  year={2026}
}

@article{curationbench,
  title={Can Generalist Agents Automate Data Curation?},
  author={Kang, Feiyang and Li, Hanze and Nguyen, Adam and Dabas, Mahavir and Ma, Jiaqi W and Sala, Frederic and Song, Dawn and Jia, Ruoxi},
  journal={arXiv preprint arXiv:2606.04261},
  year={2026}
}

@article{gspo,
  title={Group sequence policy optimization},
  author={Zheng, Chujie and Liu, Shixuan and Li, Mingze and Chen, Xiong-Hui and Yu, Bowen and Gao, Chang and Dang, Kai and Liu, Yuqiong and Men, Rui and Yang, An and others},
  journal={arXiv preprint arXiv:2507.18071},
  year={2025}
}

@article{ppo,
  title={Proximal policy optimization algorithms},
  author={Schulman, John and Wolski, Filip and Dhariwal, Prafulla and Radford, Alec and Klimov, Oleg},
  journal={arXiv preprint arXiv:1707.06347},
  year={2017}
}

@article{verl,
  title   = {HybridFlow: A Flexible and Efficient RLHF Framework},
  author  = {Guangming Sheng and Chi Zhang and Zilingfeng Ye and Xibin Wu and Wang Zhang and Ru Zhang and Yanghua Peng and Haibin Lin and Chuan Wu},
  year    = {2024},
  journal = {arXiv preprint arXiv: 2409.19256}
}

@article{unifypt,
  title={Towards a unified view of large language model post-training},
  author={Lv, Xingtai and Zuo, Yuxin and Sun, Youbang and Liu, Hongyi and Wei, Yuntian and Chen, Zhekai and Zhu, Xuekai and Zhang, Kaiyan and Wang, Bingning and Ding, Ning and others},
  journal={arXiv preprint arXiv:2509.04419},
  year={2025}
}

@article{dump,
  title={Dump: Automated distribution-level curriculum learning for rl-based llm post-training},
  author={Wang, Zhenting and Cui, Guofeng and Li, Yu-Jhe and Wan, Kun and Zhao, Wentian},
  journal={arXiv preprint arXiv:2504.09710},
  year={2025}
}

@article{asha,
  title={A system for massively parallel hyperparameter tuning},
  author={Li, Liam and Jamieson, Kevin and Rostamizadeh, Afshin and Gonina, Ekaterina and Ben-Tzur, Jonathan and Hardt, Moritz and Recht, Benjamin and Talwalkar, Ameet},
  journal={Proceedings of machine learning and systems},
  volume={2},
  pages={230--246},
  year={2020}
}

@misc{exts,
      title={Exploit More, Explore Smarter for Budget-Constrained Agentic Search}, 
      author={Haoyang Fang and Bernie Wang},
      year={2026},
      eprint={2608.23848},
      archivePrefix={arXiv},
      primaryClass={cs.AI},
      url={https://arxiv.org/abs/2608.23848}, 
}

\newpage
\tableofcontents
\newpage


\appendix

\section{Related Work}
\label{sec:related}

\paragraph{Multi-stage RL post-training.}
Progressive response length extension is the dominant pattern: DeepScaleR~\citep{deepscaler} introduced 3-stage length scaling (8K$\to$16K$\to$24K), replicated by AceReason-Nemotron~\citep{acereason} (4 stages to 32K), DeepCoder~\citep{deepcoder} (code, 16K$\to$32K), MiMo~\citep{mimo} (32K$\to$48K), and others~\citep{skywork_or1,jt_math,still3,qwenlong_l1,dsr1}. Data difficulty staging~\citep{acereason,tacler,fastcurl,diff_aware_staged_rl} and compression-extension cycles~\citep{fastcurl,siri} provide complementary patterns, while most of them hold learning rate, KL coefficient, temperature, and batch size constant across stages, leaving most of the hyperparameter space unexplored at transitions.

\paragraph{Adaptive training and HPO.}
Population-based training (PBT)~\citep{pbt} is the closest methodological ancestor: it also discovers hyperparameter schedules during training by evolving a population of configurations. The key differences are: (1)~PBT transitions via scalar fitness ranking and random perturbation, while \ours{} diagnoses specific pathologies and proposes coordinated interventions; (2)~PBT requires a population (typically 10--80 parallel workers), while \ours{} operates sequentially with checkpoint reuse under severe budget constraints; (3)~PBT's output is an opaque schedule, while \ours{}'s diagnostic analysis produces \revised{interpretable, human-readable diagnoses and a candidate design heuristic (\S\ref{sec:patterns}), scoped to our evaluated setting}. PBT excels in GPU-rich regimes where broad parallel exploration is cheap; even a minimal population of 8 workers would require 256 concurrent A100 GPUs (8$\times$32 GPUs per training run), placing it beyond our compute budget. Random search~\citep{randomhpo}, Bayesian optimization~\citep{bayesianhpo}, and Hyperband~\citep{hyperpod} search over static configurations without adapting within a training run. \revised{ASHA~\citep{asha} likewise allocates budget across static configurations via early stopping. We view these methods as complementary to and composable with \ours{} rather than as pure competitors: the LLM proposer could replace PBT's random perturbation with dynamics-aware coordinated proposals, or steer ASHA's sampling toward promising regions, which is left to future work.} \ours{} searches over configuration \emph{sequences} conditioned on training dynamics, staging parameters the literature holds constant with non-monotonic trajectories.

\paragraph{Agentic ML automation.}
Several concurrent systems apply LLM agents with tree-structured exploration to ML automation. SELA~\citep{sela} uses MCTS for ML pipeline configuration; AIDE~\citep{aide} applies tree search to data science competitions; MLZero~\citep{mlzero} provides end-to-end automation across modalities; and AlphaEvolve~\citep{alphaevolve} applies evolutionary search to code. \ours{} targets a fundamentally different search space: RL post-training trajectories where (1)~each evaluation costs much more GPU time, imposing severe budget constraints; (2)~nodes are not independent solutions but training \emph{phases} that compose via checkpoint resumption into multi-stage strategies; (3)~the search must reason about non-stationary training dynamics (KL divergence spikes, model collapse, reward stagnation, etc.) rather than static metrics; and (4)~the proposer must make coordinated multi-dimensional hyperparameter changes conditioned on the observed training state. These challenges motivate our dynamics-aware proposal mechanism and agentic early stopping.

Our contributions to the search process are: redefining nodes as training phases with checkpoint composition (\S\ref{sec:checkpoint}), dynamics-aware multimodal proposals (\S\ref{sec:reflection}), agentic early stopping (\S\ref{sec:early_stopping}), and forced from-scratch injection (\S\ref{sec:checkpoint}). The UCT value function with virtual child competition is adopted from prior work. Tree-of-Thought~\citep{tot} and RAP~\citep{hao2023reasoning} apply tree search at inference time where simulations are cheap; \ours{} applies it at training time where each node requires hours of GPU compute.

While our approach focuses on optimizing RL training trajectories, a concurrent line of work shifts the focus of agentic ML automation toward autonomous data collection and curation. For instance, PostTrainBench~\citep{posttrainbench} evaluates agents on their ability to autonomously gather and curate external data to optimize the post-training phase of base LLMs. Similarly, the DataMaster~\citep{datamaster} framework isolates the data engineering process, using tree-structured search and cumulative memory to let agents discover, clean, and compose datasets without altering the underlying learning algorithm. Exploring a related automated curation loop, Curation-Bench~\citep{curationbench} demonstrates that properly scaffolded agents can autonomously compose highly efficient data-selection policies that outperform standard baselines. These data-centric approaches complement our dynamics-aware search by targeting dataset optimization rather than training trajectory search.

\paragraph{LLM post-training methods.}
Current LLM pipelines utilize a variety of optimization algorithms for RL training, including PPO~\citep{ppo}, DPO~\citep{dpo}, GRPO~\citep{dsmath}, GSPO~\citep{gspo}, etc. \ours{} does not introduce a new alignment objective; instead, it automates the \emph{training strategy} required to effectively deploy them. We demonstrate the efficacy of our approach specifically using GRPO, leaving its application to other algorithms for future work.

\section{Experimental Setup Details}
\label{app:setup}

\subsection{Task Details}
\label{app:datasets}

Table~\ref{tab:datasets_full} summarizes the four evaluation tasks. All datasets are uniformly subsampled to 5{,}000 train, 500 validation, and 500 test examples. The dynamic orchestration workflow automatically designs a task-specific reward function for each dataset.

\begin{table*}[h]
\caption{Evaluation tasks. All tasks train with GRPO via VeRL. For ChemCoT we report the average over subtasks in this table.}
\label{tab:datasets_full}
\centering
\small
\begin{tabular}{llllrcc}
\toprule
\textbf{Dataset} & \textbf{Domain} & \textbf{Answer format} & \textbf{Metric} & \textbf{Train} & \textbf{0.6B} & \textbf{4B} \\
\midrule
ChemCoT & Chemistry (19 subtasks) & Mixed (SMILES, MCQ) & Task-specific & 5,000 & 15.4 & 14.6 \\
PaperSearchQA & Biomedical QA & Short text & Exact match & 5,000 & 11.6 & 31.6 \\
SSMR-Bench & Music theory & MCQ (A to D) & EM accuracy & 5,000 & 32.4 & 51.4 \\
WildSci & Multi-discipline science & MCQ (A to J) & EM accuracy & 5,000 & 29.4 & 53.6 \\
\bottomrule
\end{tabular}
\end{table*}

All datasets use fixed, deterministic splits with subsampled train ($\leq$5,000), validation ($\leq$500), and test ($\leq$500) sets. Base model performance (0.6B and 4B columns) is measured via greedy decoding (pass@1, temperature=0) with extended thinking and max response length=8192 on 2 nodes $\times$ 8 A100 40G GPUs.

\subsection{Baseline Configurations}

\textbf{Practitioner baseline.} A carefully tuned general-purpose GRPO recipe for Qwen3 (on 8x A100 40G GPUs), developed through internal iteration on separate tasks (e.g., math reasoning) and applied without task-specific modification. Single static configuration ($L=1$).

\textbf{Random search.} 8 trials with configurations sampled from: learning rate $\sim$ LogUniform(1e-5, 1e-4), KL coefficient $\sim$ LogUniform(1e-5, 1e-3), temperature $\sim$ Uniform(0.6, 1.2), clip ratio $\sim$ Uniform(0.15, 0.35), batch size $\in \{$64, 128, 256$\}$, LoRA rank $\in \{$16, 32, 64, 128$\}$, rollout count $\in \{$6, 8, 12$\}$, epochs $\in \{$3, 5$\}$. Each trial runs to completion without early stopping.

\textbf{Skill-based LLM agent.} An autonomous LLM agent (Claude Opus 4.6) built on the same training infrastructure to \ours{} (VeRL, reward functions, hyperparameter space). Unlike \ours{}'s fixed workflow where each stage has a deterministic prompt template, the iterative agent uses a skill-based general workflow: the LLM autonomously plans and orchestrates each iteration, invoking skills for dataset preparation, reward design, job submission, metric diagnosis, and checkpoint resume. This generality makes it more flexible but less token-efficient (\revised{73$\times$} higher API cost) since the LLM must maintain full context across all decisions. It can stop training at any time and resume from any previous checkpoint, without explicit tree search or visual reasoning.

\subsection{Compute Budget Fairness}
\label{app:budget_fairness}

Random and grid search each run 8 full-length training runs to completion and achieve similar total GPU-hours to \ours{}. Both \ours{} and the skill-based LLM agent have a maximum of 16 iterations, but the skill-based agent is limited to less than 700 GPU-hours of total runtime due to its significant API cost (Table~\ref{tab:api_cost}). This budget is sufficient for convergence: the agent does not force restarts from scratch, and its validation scores begin to plateau or decrease near the end of its time limit (Table~\ref{tab:agent_results}, Figure~\ref{fig:compute}). Notably, \ours{}'s best discovered strategy on all four tasks originates from the root node (the practitioner default configuration), indicating that the default is a strong starting point and that forced from-scratch injection, while necessary for robustness, was not the source of the best strategies in these experiments.

\subsection{API Cost}
\label{app:api_cost}

Table~\ref{tab:api_cost} compares API costs between \ours{} and the skill-based LLM agent. The fixed workflow with structured prompts at each stage consumes 73$\times$ less total API cost than the agent's autonomous orchestration, which must maintain full conversation context across all decisions.

\begin{table}[h]
\caption{LLM API cost comparison (ratios from unrounded costs). \ours{}'s structured pipeline uses \revised{73$\times$} less API cost than the skill-based LLM agent.}
\label{tab:api_cost}
\centering
\scalebox{0.85}{%
\small
\begin{tabular}{lrrr}
\toprule
\textbf{Dataset} & \textbf{\ours{} (\$)} & \textbf{LLM Agent (\$)} & \textbf{Ratio} \\
\midrule
ChemCoT & 14 & 639 & 44$\times$ \\
PaperSearchQA & 11 & 635 & 58$\times$ \\
WildSci & 15 & 1{,}125 & 74$\times$ \\
SSMR-Bench & 8 & 1{,}146 & 144$\times$ \\
\midrule
\textbf{Total} & \textbf{48} & \textbf{3{,}545} & \textbf{73$\times$} \\
\bottomrule
\end{tabular}%
}
\end{table}

\subsection{Model and Infrastructure}
All experiments use Qwen3-4B as the base model with LoRA. Training uses a modified version of VeRL (for better LoRA support, etc.) on Ray clusters deployed on AWS EKS with 4 to 8$\times$A100 40G nodes. The scaling experiment (Table~\ref{tab:scaling}) additionally evaluates Qwen3-0.6B, 1.7B, and 8B.

\subsection{\ours{} Configuration}
$C = 1.414$, $T = 0.3$, $w_f = 0.5$, $w_e = 0.3$, $o_f = 2$, max evolve children = 2, max debug children = 3, max debug depth = 5, $\rho_{\min} = 0.2$, early stopping interval = 900s. \revised{LLM backbone (main results): Claude Opus 4.5 for the proposer and the early stopper; the skill-based LLM agent uses Claude Opus 4.6 (Appendix~\ref{app:skill_agent}). The framework is backbone-agnostic, but stronger reasoning models are preferred (Appendix~\ref{app:backbone}). The additional public-stack experiments in Appendix~\ref{app:public_stack} use Claude Opus 4.8 as the default backbone.}

\subsection{Skill-Based LLM Agent}
\label{app:skill_agent}

The skill-based LLM agent uses Claude Opus 4.6 as an autonomous orchestrator with unrestricted tool access. Rather than a fixed pipeline, the agent is equipped with 8 \emph{skills}, which are natural language instruction sets that the LLM invokes as needed:

\begin{enumerate}
    \item \texttt{prepare-dataset}: Download and convert data to VeRL parquet format
    \item \texttt{define-reward}: Analyze dataset and write a \texttt{compute\_score()} reward function
    \item \texttt{validate-run}: Pre-flight evaluation on a sample
    \item \texttt{generate-config}: Write a full VeRL sweep configuration
    \item \texttt{download-model}: Download the base model
    \item \texttt{submit-training}: Generate and submit SLURM jobs
    \item \texttt{check-training}: Monitor metrics, diagnose training health, recommend action
    \item \texttt{gather-results}: Parse outputs and write a report
\end{enumerate}

The agent's optimization loop proceeds as: (1)~submit a training job, (2)~poll metrics at 5--30 minute intervals, (3)~perform a 14-parameter diagnostic analysis producing a per-parameter verdict (KEEP/INCREASE/DECREASE) based on observed KL divergence, gradient norms, clip ratios, and validation trends, (4)~decide whether to continue, early-stop, tune-and-resume from the best checkpoint, or restart from scratch.

The agent can also be coupled with \ours{} to form an end-to-end training experience with static workflow as search backend.

\paragraph{Key differences from \ours{}.} The agent maintains full conversation context across all decisions (leading to long token histories), plans its own workflow (may skip or reorder stages), and tends to always resume from the single best checkpoint in a linear chain. It analyzes only numerical metrics (no visual training curves). These design choices make it more flexible than \ours{}'s fixed harness but \revised{73$\times$} less token-efficient.

\paragraph{Budget limitation.} Due to high API cost (\$635--1{,}146 per task), the agent was limited to 6--9 training iterations (488--709 GPU-hours) rather than the full 16-iteration budget allocated to \ours{}. This limitation is inherent to the agent's design: autonomous orchestration requires maintaining full conversation context, making it \revised{73$\times$} more expensive in API cost overall than \ours{}'s fixed prompts. A budget-matched comparison at equal GPU-hours would require either reducing the agent to 1--2 iterations or spending \$10K+ per task, neither of which yields a meaningful evaluation. Despite this handicap, it achieves competitive results (Table~\ref{tab:main_results}), confirming that LLM reasoning about training dynamics is broadly effective. The advantage of \ours{} comes from its tree structure and fixed harness rather than from a fundamentally different reasoning capability.

\section{Additional Results and Analysis}
\label{app:results}

\subsection{Search Convergence}
\label{app:convergence}

\begin{figure}[h]
    \centering
    \begin{subfigure}[t]{0.48\columnwidth}
        \centering
        \includegraphics[width=\textwidth]{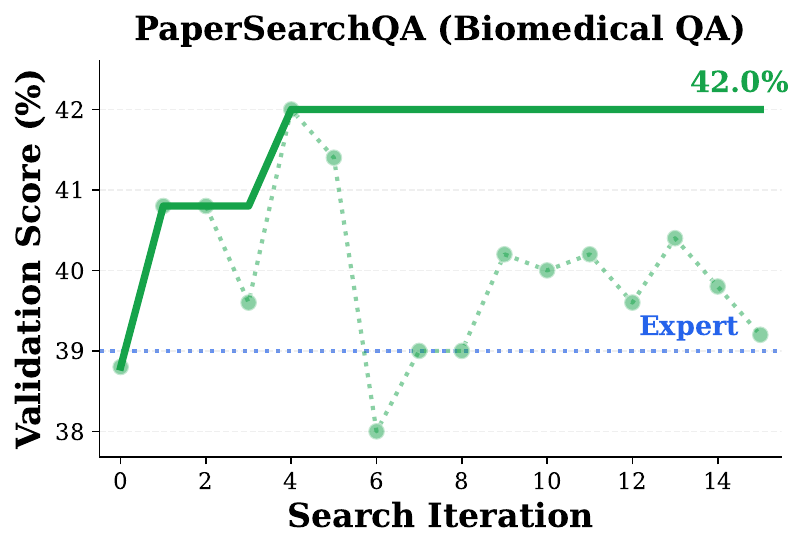}
    \end{subfigure}
    \hfill
    \begin{subfigure}[t]{0.48\columnwidth}
        \centering
        \includegraphics[width=\textwidth]{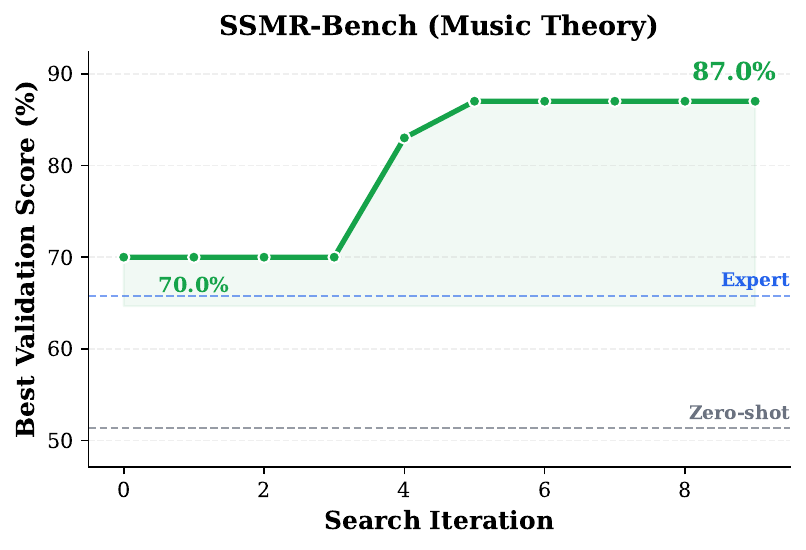}
    \end{subfigure}
    \\[0.3em]
    \begin{subfigure}[t]{0.48\columnwidth}
        \centering
        \includegraphics[width=\textwidth]{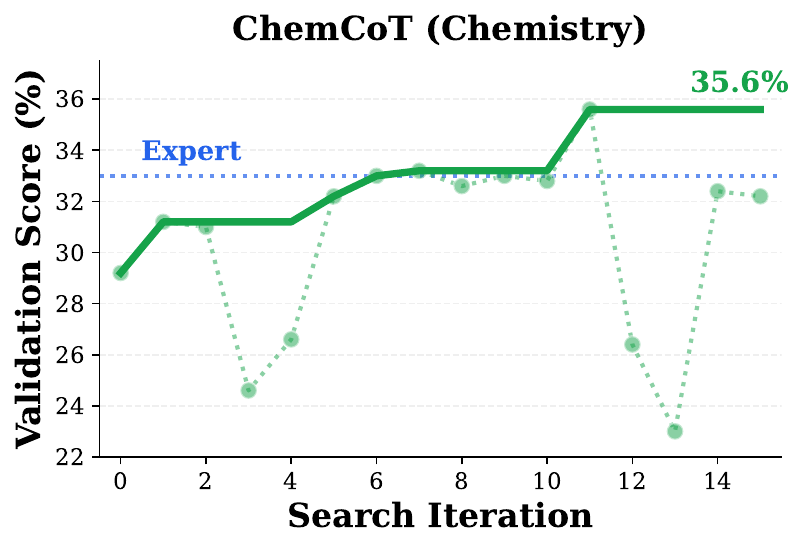}
    \end{subfigure}
    \hfill
    \begin{subfigure}[t]{0.48\columnwidth}
        \centering
        \includegraphics[width=\textwidth]{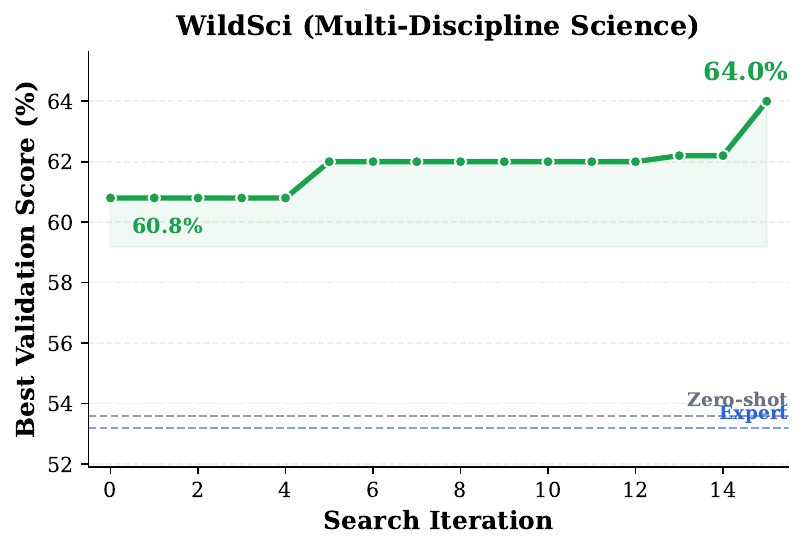}
    \end{subfigure}
    \caption{Best-so-far validation score vs.\ search iteration. \ours{} surpasses the practitioner baseline early in the search and continues improving throughout the budget.}
    \label{fig:convergence}
\end{figure}

Figure~\ref{fig:convergence} shows best-so-far validation score vs.\ search iteration for all 4 tasks. \ours{} surpasses the practitioner baseline early in the search and continues improving throughout the budget.

\subsection{ChemCoT Per-Subtask Breakdown}
\label{app:chemcot_subtasks}

Table~\ref{tab:chemcot_subtasks} reports per-subtask test accuracy for ChemCoT.

\begin{table*}[t]
\caption{ChemCoT per-subtask test accuracy (\%) for Qwen3-4B (pass@1, rl=8192), the first training run (Node~0), and the best discovered strategy (Node~11, 5-phase adaptive). Subtasks sorted by Node~11 accuracy. Nine subtasks scoring 0\% across all conditions are omitted (all \texttt{mol\_opt\_*}, \texttt{rxn\_retro}, \texttt{rxn\_nepp}, \texttt{rxn\_mechanism}). \textit{Note that the average number here refers to average over subtasks which is different from the average over domains in main table.}}
\label{tab:chemcot_subtasks}
\centering
\small
\begin{tabular}{llccc}
\toprule
\textbf{Category} & \textbf{Subtask} & \textbf{Base} & \textbf{Node 0} & \textbf{Node 11} \\
\midrule
\multirow{5}{*}{Mol.\ Understanding}
 & fg\_count       & 75.9 & 82.8\gain{6.9}  & \textbf{96.6}\gain{20.7} \\
 & ring\_count     & 16.1 & 80.6\gain{64.5} & \textbf{93.5}\gain{77.4} \\
 & ring\_system    & 70.0 & 83.3\gain{13.3} & \textbf{86.7}\gain{16.7} \\
 & equivalence     & 37.0 & \textbf{55.6}\gain{18.6} & \textbf{55.6}\gain{18.6} \\
 & murcko          & 3.3  & 3.3\gain{0.0}   & \textbf{16.7}\gain{13.4} \\
\midrule
\multirow{3}{*}{Mol.\ Editing}
 & substitution    & 0.0  & 31.0\gain{31.0} & \textbf{58.6}\gain{58.6} \\
 & deletion        & 20.7 & \textbf{37.9}\gain{17.2} & 34.5\gain{13.8} \\
 & addition        & 3.4  & 0.0             & \textbf{6.9}\gain{3.5} \\
\midrule
\multirow{2}{*}{Reaction}
 & reagent/catalyst & 11.5 & 72.1\gain{60.6} & \textbf{90.2}\gain{78.7} \\
 & forward synth.\ byproduct & 0.0 & \textbf{2.6}\gain{2.6} & \textbf{2.6}\gain{2.6} \\
\midrule
\multicolumn{2}{l}{\textbf{Macro-avg (all 19 subtasks)}} & 12.5 & 23.6\gain{11.1} & \textbf{28.5}\gain{16.0} \\
\bottomrule
\end{tabular}
\end{table*}

\subsection{Model Scaling Detailed Results}
\label{app:scaling}

Table~\ref{tab:scaling} reports per-subtask results for the scaling analysis on SSMR-Bench.

\begin{table*}[t]
\caption{Model scaling analysis on SSMR-Bench: per-subtask test accuracy (\%) at the best validation node. Subscripts show gain over the base model. \textbf{Bold}: best per subtask per model.}
\label{tab:scaling}
\centering
\begin{tabular}{ll*{4}{c}c}
\toprule
\textbf{Model} & \textbf{Method} & {\scriptsize Scl} & {\scriptsize Bea} & {\scriptsize Cho} & {\scriptsize Int} & {\scriptsize Avg} \\
\midrule
\multirow{3}{*}{Qwen3-0.6B}
 & Qwen3 & 33.6 & 30.4 & 39.2 & 26.4 & 32.4 \\
 & Practitioner & 53.6 & 75.2 & 37.6 & 45.6 & 53.0 \\
 & \textbf{\ours{}} & \textbf{95.2}\gain{61.6} & \textbf{82.4}\gain{52.0} & \textbf{60.0}\gain{20.8} & \textbf{52.0}\gain{25.6} & \textbf{72.4}\gain{40.0} \\
\midrule
\multirow{3}{*}{Qwen3-1.7B}
 & Qwen3 & 27.2 & 31.2 & 36.0 & 32.0 & 31.6 \\
 & Practitioner & 56.8 & 50.4 & 50.4 & \textbf{51.2} & 52.2 \\
 & \textbf{\ours{}} & \textbf{77.6}\gain{50.4} & \textbf{69.6}\gain{38.4} & \textbf{55.2}\gain{19.2} & 47.2\gain{15.2} & \textbf{62.4}\gain{30.8} \\
\midrule
\multirow{3}{*}{Qwen3-4B}
 & Qwen3 & 60.0 & 44.8 & 50.4 & 50.4 & 51.4 \\
 & Practitioner & 76.8 & 60.0 & 65.6 & 60.8 & 65.8 \\
 & \textbf{\ours{}} & \textbf{94.4}\gain{34.4} & \textbf{81.6}\gain{36.8} & \textbf{77.6}\gain{27.2} & \textbf{75.2}\gain{24.8} & \textbf{82.2}\gain{30.8} \\
\midrule
\multirow{3}{*}{Qwen3-8B}
 & Qwen3 & 59.2 & 56.0 & 48.8 & 46.4 & 52.6 \\
 & Practitioner & \multicolumn{5}{c}{\emph{failed (OOM)}} \\
 & \textbf{\ours{}} & \textbf{96.8}\gain{37.6} & \textbf{88.0}\gain{32.0} & \textbf{79.2}\gain{30.4} & \textbf{70.4}\gain{24.0} & \textbf{83.6}\gain{31.0} \\
\bottomrule
\end{tabular}
\end{table*}

\subsection{Additional Per-Dataset Strategies}
\label{app:extra_strategies}

We describe the two per-dataset strategies deferred from \S\ref{sec:per_dataset}; full per-phase configurations for all four tasks appear in Tables~\ref{tab:strategy_chemcot}--\ref{tab:strategy_wildsci}.

\paragraph{PaperSearchQA (Biomedical QA).} The best strategy is a 4-phase chain (Figure~\ref{fig:all_strategies}, test=42.6\%, +11.0~pp over the base model). LR and temperature are progressively tightened through Phases~2--3 (val stabilizes at 40.8\%), then the proposer diagnoses stagnation (near-zero clip ratio, low gradient norms) and reverses course at Phase~4: LR doubles, temperature increases, and batch size decreases to break the plateau (val=42.0\%). The KL coefficient increases monotonically throughout ($0.001 \to 0.01$), unlike the non-monotonic trajectories on other tasks.

\paragraph{WildSci (Multi-Discipline Science).} The best strategy is a 4-phase chain (Figure~\ref{fig:all_strategies}, test=58.5\%). After aggressive LR scaling in Phase~2 triggers a KL divergence spike, the system tightens constraints in Phase~3. This intervention causes an entropy collapse. During this collapse, the validation score increases slightly, but the hidden test score drops. To recover entropy, Phase~4 relaxes the KL penalty, lowers the LR, and raises both the temperature and the clip ratio. This successfully recovers the hidden test performance, though it does not fully surpass the Phase~2 peak due to the negative effects of Phase~3. Noticeably, the validation score continued to rise even as test performance degraded during the entropy collapse, which suggests that monitoring comprehensive training dynamics is more robust than optimizing solely for the validation score.

\subsection{Best Discovered Strategy Configurations}
\label{app:strategy_catalog}

Tables~\ref{tab:strategy_chemcot}--\ref{tab:strategy_wildsci} detail the full hyperparameter configuration at each phase of the best discovered strategy for each task. For each phase, we report the node ID in the search tree, the training step range, the proposer's diagnosis that triggered the transition, and the complete configuration.

\begin{table*}[h]
\caption{Best discovered strategy for \textbf{ChemCoT} (5-phase, Node path: 0$\to$2$\to$3$\to$5$\to$11). Node~2 resumes from Node~1's checkpoint (Node~1 failed but produced a valid checkpoint). Final val=35.6\%, test=28.5\% (macro-avg 19 subtasks).}
\label{tab:strategy_chemcot}
\centering
\small
\begin{tabular}{clp{3.2cm}p{7.0cm}}
\toprule
\textbf{Phase} & \textbf{Node / Steps} & \textbf{Proposer Diagnosis} & \textbf{Configuration} \\
\midrule
1 & Node 0, steps 0--59 & (Initial run) & LR=5e-5, KL=0.001, T=1.0, clip=0.28, batch=128, rollout=8, resp=6144, epochs=1 \\
\midrule
2 & Node 2, steps 60--79 & Under-exploring: low reward variance under default constraints & LR=5e-5, KL=0.001, T=0.85, clip=0.3, adv\_norm=true \\
\midrule
3 & Node 3, steps 80--116 & KL divergence spike: 16$\times$ KL loss spike, response length inflating (1630$\to$2278 tokens) without val improvement & LR=2.5e-5, KL=0.005 (5$\times$ increase), T=0.7, clip=0.25 \\
\midrule
4 & Node 5, steps 117--119 & Response truncation: clip ratio rising (3\%$\to$8\%) indicating output capacity exhaustion & LR=2e-5, KL=0.003, resp=7168 (+1024), clip=0.28 \\
\midrule
5 & Node 11, steps 120--180 & Low gradient signal: gradient norms $<$0.001, validation flat at 32.8\% & LR=3e-5 ($\uparrow$), KL=0.005, T=0.8, rollout=12 ($\uparrow$), batch=96 ($\downarrow$), clip=0.25 \\
\bottomrule
\end{tabular}
\end{table*}

\begin{table*}[h]
\caption{Best discovered strategy for \textbf{PaperSearchQA} (4-phase, Node path: 0$\to$1$\to$2$\to$5). Final val=42.0\%, test=42.6\%.}
\label{tab:strategy_papersearchqa}
\centering
\small
\begin{tabular}{clp{3.2cm}p{7.0cm}}
\toprule
\textbf{Phase} & \textbf{Node / Steps} & \textbf{Proposer Diagnosis} & \textbf{Configuration} \\
\midrule
1 & Node 0, steps 0--99 & (Initial run) & LR=5e-5, KL=0.001, T=1.0, clip=0.28, batch=128, epochs=3 \\
\midrule
2 & Node 1, steps 100--119 & Noisy updates: high variance in reward signal, LR too aggressive for task difficulty & LR=2.5e-5, KL=0.002, T=0.9, clip=0.28, algo\_kl=0.002, epochs=5 \\
\midrule
3 & Node 2, steps 120--139 & Continued noise: val improvement slowing, further constraint needed & LR=1e-5 ($\downarrow$), KL=0.005 ($\uparrow$), T=0.7 ($\downarrow$), clip=0.25, algo\_kl=0.005, epochs=6 \\
\midrule
4 & Node 5, steps 140--312 & Validation plateau: val flat at 40.8\%, near-zero clip ratio, low gradient norms & LR=2e-5 ($\uparrow$), KL=0.01 ($\uparrow$), T=0.8 ($\uparrow$), clip=0.3 ($\uparrow$), batch=96 ($\downarrow$), algo\_kl=0.01 \\
\bottomrule
\end{tabular}
\end{table*}

\begin{table*}[h]
\caption{Best discovered strategy for \textbf{SSMR-Bench} (4-phase, Node path: 0$\to$1$\to$4$\to$5). Final val=87.0\%, test=82.2\%.}
\label{tab:strategy_ssmr}
\centering
\small
\begin{tabular}{clp{3.2cm}p{7.0cm}}
\toprule
\textbf{Phase} & \textbf{Node / Steps} & \textbf{Proposer Diagnosis} & \textbf{Configuration} \\
\midrule
1 & Node 0, steps 0--19 & (Initial run) & LR=5e-5, KL=0.001, clip=0.28, T=1.0, resp=6144, epochs=1 \\
\midrule
2 & Node 1, steps 20--39 & Fast initial convergence: model learning MCQ pattern quickly, apply conservative tuning & LR=3e-5 ($\downarrow$), KL=0.002 ($\uparrow$), clip=0.25, grad\_clip=0.5, adv\_norm=true, T=1.1, epochs=2 \\
\midrule
3 & Node 4, steps 40--139 & Validation regression: val regressed 70.0\%$\to$67.2\% under conservative hyperparameters & LR=5e-5 (restored), KL=0.001 (relaxed), clip=0.30, T=1.1, epochs=4 \\
\midrule
4 & Node 5, steps 140--200 & Convergence stabilization: val jumped to 83.0\%, now fine-tune with expanded capacity & LR=3e-5 ($\downarrow$), KL=0.002 ($\uparrow$), resp=7168 (+1024), T=1.0, clip=0.25, epochs=6 \\
\bottomrule
\end{tabular}
\end{table*}

\begin{table*}[h]
\caption{Best discovered strategy for \textbf{WildSci} (4-phase, Node path: 0$\to$5$\to$13$\to$15). Final val=64.0\%, test=58.5\%.}
\label{tab:strategy_wildsci}
\centering
\small
\begin{tabular}{clp{3.2cm}p{7.0cm}}
\toprule
\textbf{Phase} & \textbf{Node / Steps} & \textbf{Proposer Diagnosis} & \textbf{Configuration} \\
\midrule
1 & Node 0, steps 0--99 & (Initial run) & LR=5e-5, KL=0.001, T=1.0, clip=0.28, resp=6144 \\
\midrule
2 & Node 5, steps 100--119 & Slow progress: multi-domain task needs stronger signal, try aggressive exploration & LR=1e-4 ($\uparrow$ 2$\times$), KL=0.002, T=0.9, clip=0.25, adv\_norm=true \\
\midrule
3 & Node 13, steps 120--139 & KL divergence rising: KL loss rising without val improvement, model collapse from aggressive LR & LR=5e-5, KL=0.005 ($\uparrow$), T=0.85, clip=0.22, algo\_kl=0.002 \\
\midrule
4 & Node 15, steps 140--195 & Model collapse: entropy at $\sim$0.2 (vs.\ $\sim$0.7--0.8 healthy), policy distribution narrowed prematurely & LR=3e-5, KL=0.002 ($\downarrow$), T=0.95 ($\uparrow$), clip=0.28 ($\uparrow$), clip\_low=0.15 \\
\bottomrule
\end{tabular}
\end{table*}

\subsection{Capacity Guidebook Analysis}
\label{app:guidebook_analysis}

Comparing the Capacity Guidebook to the full SSMR transfer (Table~\ref{tab:transfer}) isolates the contribution of regularization oscillation versus pure capacity expansion.

\paragraph{WildSci.} Both schedules converge to near-identical performance (58.3\% vs.\ 58.4\% vs.\ \ours{}'s 58.5\%), confirming that expanded reasoning capacity, rather than hyperparameter oscillation, drives improvement on this task.

\paragraph{ChemCoT.} The SSMR transfer (41.3\%) substantially outperforms the Capacity Guidebook (36.6\%), indicating that LR/KL oscillation provides +4.7~pp beyond pure capacity scaling, consistent with the KL divergence spike diagnosis in \S\ref{sec:per_dataset}.

\paragraph{PaperSearchQA.} Both fixed schedules underperform the static practitioner (37.8\% and 36.2\% vs.\ 39.0\%), confirming that this task's progressive-tightening dynamic cannot be captured by any fixed multi-stage recipe.

These results \revised{support the candidate design heuristic}: capacity parameters should accumulate monotonically, but the task-specific LR and KL trajectories that \ours{} discovers are essential for realizing the full benefit of multi-stage training.

\paragraph{Transfer caveats.} Both the SSMR transfer and the Capacity Guidebook benefit from all tasks sharing the same subsampled dataset size (5{,}000 training examples), which produces similar steps-per-epoch and possibly comparable training dynamics timelines. Transferring fixed schedules between datasets of substantially different sizes may require recalibrating transition points, as the step at which the model saturates its current response budget depends on dataset complexity and size. This is an additional advantage of adaptive methods: they discover appropriate transition points from observed dynamics regardless of dataset scale.

\section{Further Analysis of the Main Results}
\label{app:main_analysis}

\revised{
The analyses below further examine our main-paper results (internal VeRL build; Qwen3-4B, except the 8B recovery, which comes from the scaling analysis of \S\ref{sec:scaling}). They introduce no new training runs and are separate from the public VeRL 0.7.1 suite of Appendix~\ref{app:public_stack}.
}

\subsection{Matched-Budget Comparison with the Skill-Based Agent}
\label{app:matched_budget}
\revised{
The skill agent's 6--9 iterations follow from its 73$\times$ higher API cost (Table~\ref{tab:api_cost}), not a design choice. Capping \ours{} at its first 8 iterations (within that range) and scoring each method at its best-so-far validation node (per-node tables in Appendix~\ref{app:detailed_results}), \ours{} matches or exceeds the skill agent on all four tasks (Table~\ref{tab:matched_budget}); on ChemCoT it still beats the skill agent (37.4 vs.\ 37.0) but ranks second behind random search. The iteration gap thus does not explain its advantage. We keep the API-cost limit explicit rather than claim a GPU-matched win (a GPU-matched skill run would exceed \$10{,}000/task in API); the skill agent shares our proposer implementation and is part of our contribution, not a fully external baseline.
}

\begin{table}[h]
\caption{\revised{Matched refinement-iteration budget: test score (\%) at the best-so-far validation node, with \ours{} capped at its first 8 search iterations (within the skill agent's 6--9 range). WildSci here reads 58.6 from the per-node table; cf.\ 58.5 for the full 16-iteration run in Table~\ref{tab:main_results}.}}
\label{tab:matched_budget}
\centering
\revised{
\scalebox{0.8}{%
\small
\begin{tabular}{lcc}
\toprule
\textbf{Task} & \textbf{Skill-based agent} & \textbf{\ours{} (first 8 iters)} \\
\midrule
ChemCoT       & 37.0 & 37.4 \\
PaperSearchQA & 40.2 & 42.6 \\
SSMR-Bench    & 80.0 & 82.2 \\
WildSci       & 56.6 & 58.6 \\
\bottomrule
\end{tabular}%
}%
}
\end{table}

\subsection{Detailed 8B Out-of-Memory Recovery}
\label{app:8b_oom}
\revised{
From the run logs (Qwen3-8B, SSMR-Bench): the practitioner default failed at vLLM rollout-engine initialization, before any training step (\emph{``No available memory for the cache blocks\ldots''}). With \texttt{gpu\_memory\_utilization}${=}0.4$ and \texttt{tensor\_model\_parallel\_size}${=}1$, the 8B weights, FSDP actor/reference parameters, and optimizer state left too little memory for vLLM's KV-cache at the 10{,}240-token max sequence (4{,}096 prompt + 6{,}144 response).

\ours{}'s error analyzer diagnosed this and recovered autonomously by raising \texttt{gpu\_memory\_utilization} to 0.6 and \texttt{tensor\_model\_parallel\_size} to 2 and enabling \texttt{enforce\_eager}, plus two secondary fixes (raising \texttt{max\_num\_batched\_tokens} to 10{,}240 under chunked prefill, and reverting an FSDP2/DTensor incompatibility from parameter offloading), so the fix required reasoning about the KV-cache and FSDP memory layout, not a single toggle. The best 8B configuration reached 83.6\% test versus 52.6\% for the base model (Table~\ref{tab:8b_breakdown}, matching Table~\ref{tab:scaling}). As in \S\ref{sec:scaling}, we do not claim OOM recovery as a fundamental capability (a practitioner could fix it manually), only that \ours{} navigated it autonomously.
}

\begin{table}[h]
\caption{\revised{Best discovered Qwen3-8B configuration on SSMR-Bench (per-subtask test \%).}}
\label{tab:8b_breakdown}
\centering
\revised{
\scalebox{0.8}{%
\small
\begin{tabular}{lc}
\toprule
\textbf{Subtask (SSMR-Bench, Qwen3-8B)} & \textbf{\ours{} best config} \\
\midrule
Scale    & 96.8 \\
Beat     & 88.0 \\
Chord    & 79.2 \\
Interval & 70.4 \\
\midrule
Test average & 83.6 \\
\bottomrule
\end{tabular}%
}%
}
\end{table}

\subsection{Proposer Reasoning Trace}
\label{app:reasoning_trace}
\revised{
This SSMR-Bench trace shows coordinated, diagnosis-conditioned reasoning that random search cannot replicate. At Node~1 the proposer tightened five parameters at once (lower LR, doubled KL, narrower \texttt{clip\_ratio\_high}, advantage normalization, halved \texttt{grad\_clip}) to stabilize near the optimum; this over-constrained learning and regressed validation from 70.0\% to 67.2\%. At Node~4 it diagnosed and reversed that decision:

\begin{quote}\itshape
``The training shows clear signs of plateauing with validation score stuck at 0.67 since step 25, while the best historical run achieved 0.70. The KL loss is steadily increasing (0.001 -> 0.012) indicating policy drift from the reference, but this isn't translating to validation gains. The clip ratio decreased significantly (0.11 -> 0.04), suggesting the policy updates became more conservative over time... The gradient norms remain stable (\textasciitilde{}0.0014--0.0019), so training mechanics are healthy but the learning signal appears exhausted.''
\end{quote}

It then loosened five parameters together, each with a causal justification: raising LR (low gradient norms plus a plateau imply too-conservative steps), widening \texttt{clip\_ratio\_high} (updates had grown too conservative), raising temperature (to counter entropy decline), and lowering both the KL loss and reward-side KL coefficients (the penalty was over-constraining useful updates), driving validation from 67.2\% to 83.0\%. Random search samples configurations independently and cannot condition such a coordinated loosening on a specific diagnosis, nor reverse a specific earlier over-constraint.
}

\subsection{Failure Modes Beyond Compute Limits}
\label{app:failure_modes}
\revised{
Beyond compute limits, \ours{} has two failure modes. (1)~The proposer can misinterpret a hyperparameter, metric, or dynamics signal when the grounding knowledge is absent, which is why we inject knowledge (Appendix~\ref{app:knowledge}); e.g., LLMs routinely confuse PPO policy clipping with the response-length truncation ratio (\texttt{response\_length/clip\_ratio}), so on a new framework or unfamiliar metrics high-quality descriptions are essential. (2)~A base-model capability boundary caps what any schedule can reach: on ChemCoT, nine molecular-optimization and reaction subtasks score 0\% for every method, as valid SMILES generation is a capability the 4B base model lacks (\S\ref{sec:per_dataset}). Adaptive scheduling optimizes \emph{how} to train but cannot elicit a rollout the model cannot produce.
}

\section{Additional Experiments on the Public VeRL 0.7.1 Stack}
\label{app:public_stack}

\revised{
All experiments here use the fully public VeRL 0.7.1 release, so they also re-confirm our main claims on a public stack, independent of the internal build. Setup: Qwen3-0.6B (LoRA rank 64), SSMR-Bench subsampled to 2{,}500 / 200 / 200 (train/val/test), 16 \ours{} MCTS iterations on one 8-GPU node, Claude Opus 4.8 backbone; we report test accuracy (\%) at the best-validation node, averaged over the four SSMR-Bench subtasks (Scl=scale, Bea=beat, Cho=chord, Int=interval). The full system reaches 61.5\% versus 44.0\% for default GRPO (4 epochs, no search) and 30.5\% for the base model (Table~\ref{tab:public_stack}). \emph{These runs use a smaller dataset (and model) than the main experiments and serve as an extended ablation, so their absolute numbers are not directly comparable to the main results.}
}

\begin{table}[h]
\caption{\revised{Ablation and backbone suite on the public VeRL 0.7.1 stack (Qwen3-0.6B, SSMR-Bench). Test accuracy (\%) at the best-validation node, averaged over four subtasks; Val is the aggregate validation score at that node.}}
\label{tab:public_stack}
\centering
\revised{
\scalebox{0.80}{%
\small
\begin{tabular}{lcccccc}
\toprule
\textbf{Condition} & \textbf{Test} & {\scriptsize Scl} & {\scriptsize Bea} & {\scriptsize Cho} & {\scriptsize Int} & {\scriptsize Val} \\
\midrule
Base model (untrained) & 30.5 & 34 & 26 & 30 & 32 & $\sim$25 \\
Default GRPO, 4 ep.\ (no search) & 44.0 & 66 & 58 & 34 & 18 & 52.0 \\
\ours{} full system (Opus 4.8) & \textbf{61.5} & 78 & 78 & 48 & 42 & 73.5 \\
\midrule
Ablation: reward-only signal & 49.5 & 64 & 66 & 42 & 26 & 56.0 \\
Backbone: Sonnet 4.6 & 55.5 & 84 & 64 & 42 & 32 & 62.5 \\
Backbone: Haiku 4.5 & 39.5 & 54 & 48 & 40 & 16 & 49.0 \\
\bottomrule
\end{tabular}%
}
}
\end{table}

\subsection{Reward-Only Reasoning Ablation}
\label{app:reward_only}
\revised{
To isolate the LLM's reasoning from the tree-search machinery, we give the proposer and early stopper only the scalar reward, removing every diagnostic signal they normally read (KL, entropy, gradient norms, clip ratios, response-length dynamics) while holding all else fixed. Test mean drops from 61.5\% to 49.5\% (Table~\ref{tab:public_stack}), a 12-point loss. Since removing the proposer entirely would collapse the system into random/grid search, this is the cleanest way to isolate the reasoning content; it complements the component-level ablations of Table~\ref{tab:ablation}.
}

\subsection{Backbone-Quality Ablation}
\label{app:backbone}
\revised{
Varying only the proposer/early-stopper backbone (all else fixed), test accuracy rises monotonically with reasoning capability: Haiku 4.5 39.5\%, Sonnet 4.6 55.5\%, Opus 4.8 61.5\% (Table~\ref{tab:public_stack}). The weakest backbone falls below the no-search default GRPO (44.0\%) while stronger ones clear it widely; were the budget alone responsible, a weaker reasoner would not regress to the static-training band. Reasoning quality, not the tree machinery alone, is thus a primary driver of search quality. The framework is backbone-agnostic but favors stronger reasoners.
}

\subsection{Poor-Default From-Scratch Ablation}
\label{app:from_scratch}
\revised{
From-scratch injection (\S\ref{sec:checkpoint}) is regime-dependent: it looks inert when the default is already strong, as on all four tasks, where the best strategy comes from the root subtree. Under a deliberately weak default ($\mathrm{KL}{=}10^{-2}$), the best node reaches 56.1\% test and descends from an injected from-scratch node, versus 51.5\% for the best root-subtree descendant (a 4.6-point gap); recovery takes several iterations as the proposer walks $10^{-2}\to5\times10^{-3}\to2\times10^{-3}$ to a healthy range. This positions the mechanism as robustness insurance, with $\rho_{\min}$ a tunable exploration knob.
}

\section{Search Algorithm Details}
\label{app:mcts}

While the UCT computation and virtual child competition mechanisms are adopted from prior work (reproduced here for completeness), we introduce the failure subtree pruning (\S\ref{sec:search}) to improve search efficiency and forced from-scratch injection mechanism (\S\ref{sec:checkpoint}) to maintain exploration diversity. As search optimization is not the primary focus of this paper, we leave the design of more sophisticated search algorithms\revised{ that could further improve performance}, particularly those tailored for budget constraints\revised{~\citep{exts}} or asynchronous execution, to future work.

\subsection{Search Loop Pseudocode}

Algorithm~\ref{alg:mcts} summarizes the main loop. At each iteration, UCT selection traverses the tree to choose a parent node for expansion. If the selected parent failed, an error analyzer diagnoses the failure and proposes a fix; otherwise, the proposer agent analyzes the parent's training dynamics and proposes a new configuration with a checkpoint to resume from. The resulting child node is executed as a training job, monitored by the agentic early stopper at fixed intervals. Upon completion (or early termination), the best validation score is backpropagated up the tree and terminal subtrees are pruned. The final output is the scratch-to-leaf path with the highest validation score, interpreted as a multi-stage adaptive strategy.

\begin{algorithm}[t]
\caption{\ours{} adaptive strategy discovery loop}
\label{alg:mcts}
\begin{algorithmic}[1]
\REQUIRE Dataset $\mathcal{D}$, base model $M_0$, budget $B$
\STATE Initialize root node $n_0$ with default configuration $\theta_0$
\STATE Run initialization agents (data perception, task descriptor, tool selector)
\FOR{$t = 1$ to $B$}
    \STATE $n_{\text{parent}} \gets$ \textsc{SelectNode}(tree) \COMMENT{UCT selection (Appendix~\ref{app:uct})}
    \IF{$n_{\text{parent}}$ failed}
        \STATE $(\theta_{\text{new}}, \text{fix}) \gets$ \textsc{ErrorAnalyzer}($n_{\text{parent}}$) \COMMENT{Diagnose and propose fix}
    \ELSE
        \STATE $(\theta_{\text{new}}, k) \gets$ \textsc{Proposer}($n_{\text{parent}}$.\text{metrics}, $n_{\text{parent}}$.\text{plots}) \COMMENT{Multimodal analysis}
    \ENDIF
    \STATE $n_{\text{new}} \gets$ \textsc{CreateChild}($n_{\text{parent}}$, $\theta_{\text{new}}$, checkpoint step $k$)
    \STATE Generate and submit training code via multi-agent pipeline (\S\ref{sec:system})
    \WHILE{training in progress}
        \STATE Sample metrics and generate comparison plots at intervals
        \IF{\textsc{EarlyStopper}(current metrics, best strategy plots) = \texttt{STOP}}
            \STATE Terminate training early
        \ENDIF
    \ENDWHILE
    \STATE $s \gets$ best validation score observed during run
    \STATE \textsc{Backpropagate}($n_{\text{new}}$, $s$); \textsc{PruneTerminal}($n_{\text{new}}$)
\ENDFOR
\RETURN Strategy (scratch-to-leaf path) with highest validation score
\end{algorithmic}
\end{algorithm}

\subsection{UCT Computation}
\label{app:uct}

We integrate expansion into selection via a \emph{virtual new child} that competes against existing children at every internal node. At node $p$ with existing children $\{c_1, \ldots, c_k\}$:
\begin{align}
    \text{UCT}(c_i) &= Q(c_i) + C \sqrt{\frac{\ln N(p)}{N(c_i)}}, \label{eq:uct_child} \\
    \text{UCT}(\text{new}) &= Q_{\text{prior}}(p) + C \sqrt{\frac{\ln N(p)}{N_{\text{fair}}}}, \label{eq:uct_new}
\end{align}
where $Q_{\text{prior}}(p)$ is the parent's own normalized score (encoding the prior belief that a new child will perform similarly to its parent) and $N_{\text{fair}} = N(p) / (k+1)$ gives the virtual child a fair share of the parent's visit budget.

The exploitation term $Q(c)$ uses min-max normalization followed by exponential shaping for scale invariance:
\begin{align}
    \hat{s} = \frac{s - s_{\min}}{s_{\max} - s_{\min}}, \qquad f_T(\hat{s}) = \frac{e^{\hat{s}/T} - 1}{e^{1/T} - 1},
    \label{eq:score_norm}
\end{align}
where $T$ is a temperature parameter (default 0.3). The full exploitation term combines three signals:
\begin{align}
    Q(c) &= \frac{v_{\text{val}}}{v_{\text{total}}} \cdot f_T(\bar{s}_c) \nonumber \\
    &\quad - w_f \cdot \frac{\max(0,\, v_{\text{fail}} - o_f)}{v_{\text{total}}} - w_e \cdot \frac{v_{\text{early}}}{v_{\text{total}}},
    \label{eq:exploitation}
\end{align}
where $v_{\text{val}}$, $v_{\text{fail}}$, $v_{\text{early}}$, $v_{\text{total}}$ are validated, failed, early-stopped, and total visit counts; $\bar{s}_c$ is the average normalized score of validated descendants; $w_f = 0.5$ and $w_e = 0.3$ are penalty weights; and $o_f$ (default 2) forgives the first $o_f$ failures before penalizing.

\subsection{Virtual New Child Competition}

At each internal node $p$ during selection traversal:
\begin{enumerate}
    \item Compute UCT for all existing non-terminal children $\{c_1, \ldots, c_k\}$ (Eq.~\ref{eq:uct_child}).
    \item Compute UCT for the virtual new child (Eq.~\ref{eq:uct_new}) with $Q_{\text{prior}} = f_T(\hat{s}_p)$ and $N_{\text{fair}} = N(p)/(k+1)$.
    \item If $\text{UCT}(\text{new}) > \max_i \text{UCT}(c_i)$ \textbf{and} $k < k_{\max}$: expand (create new child at $p$).
    \item Otherwise: descend into $\arg\max_i \text{UCT}(c_i)$ and repeat.
\end{enumerate}

This mechanism naturally adapts breadth vs.\ depth: when children underperform their parent, the virtual child's prior wins, triggering exploration of a new transition from the same checkpoint.

\section{Detailed Per-Run Results}
\label{app:detailed_results}

This section reports the full hyperparameter configuration and performance for every run across all methods and tasks.

\subsection{HPO Baseline Configurations}

\paragraph{Practitioner baseline.} Table~\ref{tab:practitioner_config} reports the practitioner configuration, a carefully tuned general-purpose GRPO recipe applied identically to all tasks.

\begin{table}[h]
\caption{Practitioner baseline configuration (applied to all tasks without modification).}
\label{tab:practitioner_config}
\centering
\small
\begin{tabular}{ll}
\toprule
\textbf{Parameter} & \textbf{Value} \\
\midrule
Learning rate & 5e-5 \\
KL coefficient & 0.001 \\
Temperature & 1.0 \\
Clip ratio (high) & 0.28 \\
Train batch size & 128 \\
LoRA rank / alpha & 64 / 64 \\
Rollout count ($n$) & 8 \\
Epochs & 1 \\
Response length & 6144 \\
Gradient clip & 1.0 \\
Weight decay & 0.01 \\
Advantage normalization & false \\
\bottomrule
\end{tabular}
\end{table}

\paragraph{Capacity Guidebook.} Table~\ref{tab:guidebook_config} reports the 3-phase capacity-scaling schedule. All hyperparameters not listed remain at practitioner defaults (Table~\ref{tab:practitioner_config}). Phase durations are inspired by transition points observed in \ours{}'s discovered strategies.

\begin{table}[h]
\caption{Capacity Guidebook configuration. Progressive capacity scaling with all other hyperparameters fixed to practitioner defaults (Table~\ref{tab:practitioner_config}).}
\label{tab:guidebook_config}
\centering
\small
\begin{tabular}{clcc}
\toprule
\textbf{Phase} & \textbf{Steps} & \textbf{Response Length} & \textbf{Rollout $n$} \\
\midrule
1 & 0--60   & 4096 & 6 \\
2 & 60--120 & 6144 & 8 \\
3 & 120--180 & 8192 & 12 \\
\bottomrule
\end{tabular}
\end{table}

\paragraph{Grid search.} All grid runs use the practitioner defaults except for two swept parameters: LR $\in \{$1e-5, 3e-5, 5e-5, 1e-4$\}$ $\times$ LoRA rank $\in \{$64, 128$\}$ (8 configurations per task). \textbf{Grid runs additionally use epochs=5 and response length=8192}.

\paragraph{Random search.} Each run samples all hyperparameters independently. Table~\ref{tab:hpo_random_configs} reports the sampled configurations.

\begin{table*}[h]
\caption{Random search hyperparameter configurations for all tasks. LR=learning rate, KL=KL coefficient, T=temperature, Cl=clip ratio, BS=batch size, Rk=LoRA rank, Ro=rollouts, Ep=epochs, RL=response length.}
\label{tab:hpo_random_configs}
\centering
\scalebox{0.72}{%
\small
\begin{tabular}{cl*{9}{c}}
\toprule
\textbf{Task} & \textbf{Run} & \textbf{LR} & \textbf{KL} & \textbf{T} & \textbf{Cl} & \textbf{BS} & \textbf{Rk} & \textbf{Ro} & \textbf{Ep} & \textbf{RL} \\
\midrule
\multirow{8}{*}{\rotatebox{90}{PaperSearchQA}}
& 00 & 2.65e-5 & 5.1e-5 & 0.98 & 0.17 & 64 & 64 & 6 & 3 & 8192 \\
& 01 & 5.68e-5 & 2.6e-4 & 0.61 & 0.26 & 128 & 128 & 6 & 3 & 4096 \\
& 02 & 1.69e-5 & 1.5e-4 & 0.65 & 0.30 & 256 & 16 & 12 & 3 & 8192 \\
& 03 & 1.07e-5 & 1.9e-5 & 0.96 & 0.22 & 64 & 128 & 6 & 5 & 8192 \\
& 04 & 5.32e-5 & 2.4e-4 & 0.70 & 0.24 & 64 & 32 & 8 & 5 & 8192 \\
& 05 & 8.99e-5 & 2.7e-4 & 0.66 & 0.32 & 64 & 128 & 8 & 5 & 8192 \\
& 06 & 1.06e-5 & 7.1e-4 & 0.85 & 0.21 & 64 & 32 & 6 & 5 & 8192 \\
& 07 & 6.00e-5 & 5.0e-4 & 0.83 & 0.26 & 256 & 64 & 12 & 3 & 8192 \\
\midrule
\multirow{8}{*}{\rotatebox{90}{WildSci}}
& 00 & 2.33e-5 & 3.0e-4 & 0.71 & 0.26 & 64 & 64 & 6 & 3 & 8192 \\
& 01 & 1.02e-5 & 2.0e-5 & 1.02 & 0.19 & 128 & 32 & 8 & 3 & 4096 \\
& 02 & 7.11e-5 & 1.1e-4 & 0.64 & 0.28 & 64 & 64 & 8 & 5 & 4096 \\
& 03 & 2.12e-5 & 5.4e-4 & 1.06 & 0.28 & 128 & 32 & 6 & 5 & 8192 \\
& 04 & 2.44e-5 & 5.8e-5 & 0.84 & 0.24 & 128 & 128 & 8 & 5 & 4096 \\
& 05 & 1.40e-5 & 5.8e-4 & 0.78 & 0.24 & 64 & 16 & 6 & 3 & 4096 \\
& 06 & 6.63e-5 & 2.2e-4 & 0.88 & 0.27 & 256 & 128 & 6 & 5 & 8192 \\
& 07 & 4.19e-5 & 8.1e-4 & 1.08 & 0.18 & 256 & 32 & 12 & 5 & 8192 \\
\midrule
\multirow{8}{*}{\rotatebox{90}{SSMR-Bench}}
& 00 & 6.07e-5 & 1.1e-4 & 0.80 & 0.16 & 128 & 32 & 8 & 5 & 8192 \\
& 01 & 4.25e-5 & 2.0e-5 & 0.95 & 0.26 & 64 & 128 & 8 & 3 & 8192 \\
& 02 & 6.04e-5 & 1.2e-5 & 0.84 & 0.34 & 64 & 64 & 6 & 3 & 8192 \\
& 03 & 2.85e-5 & 1.8e-4 & 0.92 & 0.22 & 128 & 64 & 8 & 5 & 8192 \\
& 04 & 3.18e-5 & 4.3e-5 & 0.85 & 0.17 & 128 & 128 & 12 & 5 & 4096 \\
& 05 & 5.13e-5 & 2.4e-4 & 0.94 & 0.33 & 256 & 16 & 12 & 3 & 4096 \\
& 06 & 1.97e-5 & 2.9e-5 & 0.68 & 0.17 & 64 & 64 & 8 & 5 & 8192 \\
& 07 & 1.18e-5 & 2.1e-4 & 0.76 & 0.31 & 256 & 64 & 8 & 3 & 8192 \\
\midrule
\multirow{8}{*}{\rotatebox{90}{ChemCoT}}
& 00 & 1.83e-5 & 3.3e-5 & 0.84 & 0.30 & 64 & 32 & 6 & 3 & 4096 \\
& 01 & 4.71e-5 & 4.9e-5 & 0.77 & 0.16 & 128 & 32 & 6 & 5 & 4096 \\
& 02 & 1.26e-5 & 9.4e-4 & 1.09 & 0.22 & 256 & 128 & 8 & 3 & 8192 \\
& 03 & 1.78e-5 & 6.3e-4 & 1.09 & 0.32 & 64 & 64 & 6 & 5 & 8192 \\
& 04 & 5.40e-5 & 8.6e-5 & 0.91 & 0.18 & 256 & 64 & 6 & 3 & 8192 \\
& 05 & 1.43e-5 & 6.9e-5 & 1.03 & 0.23 & 64 & 16 & 6 & 5 & 4096 \\
& 06 & 7.83e-5 & 1.7e-4 & 1.10 & 0.19 & 64 & 16 & 6 & 5 & 4096 \\
& 07 & 1.31e-5 & 4.1e-5 & 1.12 & 0.20 & 128 & 16 & 12 & 5 & 4096 \\
\bottomrule
\end{tabular}%
}
\end{table*}

\subsection{Per-Run Results: Random Search}

Table~\ref{tab:random_results} reports results for each random search trial.

\begin{table*}[h]
\caption{Random search per-run results. GPU-hrs on 32$\times$A100. Best Val = best validation score achieved. Test reported at best validation step. \textbf{Bold}: best per task.}
\label{tab:random_results}
\centering
\scalebox{0.82}{%
\small
\begin{tabular}{cl*{4}{c}}
\toprule
\textbf{Task} & \textbf{Run} & \textbf{GPU-hrs} & \textbf{Best Val} & \textbf{Best Step} & \textbf{Test} \\
\midrule
\multirow{8}{*}{PaperSearchQA}
& 00 & 470.6 & 0.370 & 234 & 0.360 \\
& 01 & 171.1 & \textbf{0.392} & 93 & 0.376 \\
& 02 & 198.5 & 0.370 & 36 & 0.366 \\
& 03 & 1740.6 & 0.384 & 84 & 0.358 \\
& 04 & 837.8 & 0.386 & 198 & 0.356 \\
& 05 & 1081.9 & 0.384 & 165 & 0.354 \\
& 06 & 510.8 & 0.384 & 198 & 0.364 \\
& 07 & 191.7 & 0.390 & 45 & \textbf{0.384} \\
\midrule
\multirow{8}{*}{WildSci}
& 00 & 602.3 & 0.604 & 147 & 0.565 \\
& 01 & 240.8 & 0.600 & 72 & 0.555 \\
& 02 & 577.7 & 0.612 & 72 & 0.559 \\
& 03 & 580.1 & 0.628 & 138 & \textbf{0.580} \\
& 04 & 1137.4 & \textbf{0.630} & 177 & 0.558 \\
& 05 & 295.6 & 0.610 & 132 & 0.544 \\
& 06 & 451.0 & \textbf{0.630} & 66 & 0.578 \\
& 07 & 806.7 & 0.602 & 27 & 0.540 \\
\midrule
\multirow{8}{*}{SSMR-Bench}
& 00 & 2193.9 & 0.708 & 186 & 0.682 \\
& 01 & 779.5 & 0.696 & 120 & 0.614 \\
& 02 & 575.1 & 0.754 & 213 & 0.696 \\
& 03 & 1536.0 & --- & --- & --- \\
& 04 & 551.3 & 0.758 & 189 & 0.738 \\
& 05 & 279.8 & \textbf{0.778} & 57 & \textbf{0.744} \\
& 06 & 483.1 & 0.550 & 36 & 0.516 \\
& 07 & 431.6 & 0.688 & 57 & 0.660 \\
\midrule
\multirow{8}{*}{ChemCoT}
& 00 & 324.1 & 0.288 & 189 & 0.334 \\
& 01 & 1095.5 & 0.288 & 165 & 0.330 \\
& 02 & 356.9 & 0.270 & 30 & 0.297 \\
& 03 & 1015.2 & 0.304 & 255 & 0.316 \\
& 04 & 258.0 & 0.272 & 27 & 0.280 \\
& 05 & 504.4 & 0.282 & 228 & 0.324 \\
& 06 & 461.5 & \textbf{0.320} & 354 & \textbf{0.394} \\
& 07 & 527.1 & 0.290 & 174 & 0.313 \\
\bottomrule
\end{tabular}%
}
\end{table*}

\subsection{Per-Run Results: Grid Search}

Table~\ref{tab:grid_results} reports results for each grid search trial.

\begin{table*}[h]
\caption{Grid search per-run results. Grid varies LR $\in \{$1e-5, 3e-5, 5e-5, 1e-4$\}$ $\times$ LoRA rank $\in \{$64, 128$\}$, while SSMR-Bench uses LR $\in \{$1e-5, 3e-5, 5e-5, 1e-4$\}$ $\times$ LoRA rank $\in \{$16, 32, 64, 128$\}$ to explore broader search space; all other HPs at default. \textbf{Bold}: best per task.}
\label{tab:grid_results}
\centering
\scalebox{0.82}{%
\small
\begin{tabular}{cl*{5}{c}}
\toprule
\textbf{Task} & \textbf{Run} & \textbf{LR} & \textbf{Rank} & \textbf{GPU-hrs} & \textbf{Best Val} & \textbf{Test} \\
\midrule
\multirow{8}{*}{PaperSearchQA}
& 08 & 1e-5 & 64 & 372.6 & 0.398 & 0.390 \\
& 09 & 3e-5 & 64 & 383.5 & 0.384 & 0.390 \\
& 10 & 5e-5 & 64 & 387.8 & 0.394 & 0.388 \\
& 11 & 1e-4 & 64 & 397.8 & 0.396 & \textbf{0.396} \\
& 12 & 1e-5 & 128 & 424.4 & 0.392 & 0.384 \\
& 13 & 3e-5 & 128 & 1230.5 & 0.388 & 0.372 \\
& 14 & 5e-5 & 128 & 411.4 & 0.386 & 0.380 \\
& 15 & 1e-4 & 128 & 1299.3 & \textbf{0.408} & 0.372 \\
\midrule
\multirow{8}{*}{WildSci}
& 08 & 1e-5 & 64 & 711.2 & 0.616 & 0.551 \\
& 09 & 3e-5 & 64 & 2112.0 & 0.622 & 0.568 \\
& 10 & 5e-5 & 64 & 714.0 & 0.626 & 0.572 \\
& 11 & 1e-4 & 64 & 715.1 & \textbf{0.650} & 0.530 \\
& 12 & 1e-5 & 128 & 763.9 & 0.620 & 0.575 \\
& 13 & 3e-5 & 128 & 762.5 & 0.632 & \textbf{0.580} \\
& 14 & 5e-5 & 128 & 1545.3 & 0.626 & 0.558 \\
& 15 & 1e-4 & 128 & 770.5 & 0.632 & 0.576 \\
\midrule
\multirow{8}{*}{ChemCoT}
& 08 & 1e-5 & 64 & 684.5 & 0.296 & 0.327 \\
& 09 & 3e-5 & 64 & 703.2 & 0.308 & 0.350 \\
& 10 & 5e-5 & 64 & 725.5 & 0.304 & 0.351 \\
& 11 & 1e-4 & 64 & 1491.0 & 0.302 & 0.377 \\
& 12 & 1e-5 & 128 & 759.4 & 0.302 & 0.314 \\
& 13 & 3e-5 & 128 & 2038.7 & 0.318 & \textbf{0.381} \\
& 14 & 5e-5 & 128 & 766.4 & \textbf{0.322} & 0.350 \\
& 15 & 1e-4 & 128 & 2062.4 & 0.320 & 0.363 \\
\bottomrule
\end{tabular}%
}
\end{table*}

\subsection{Per-Iteration Results: Skill-Based LLM Agent}

Table~\ref{tab:agent_results} reports per-iteration results for the skill-based LLM agent.

\begin{table*}[h]
\caption{Skill-based LLM agent per-iteration results. GPU-hrs reported per iteration. Budget-limited due to API cost (\$635--1,146 per task). Best Val = best validation score at any step within the iteration.}
\label{tab:agent_results}
\centering
\scalebox{0.82}{%
\small
\begin{tabular}{cl*{5}{c}}
\toprule
\textbf{Task} & \textbf{Iter} & \textbf{GPU-hrs} & \textbf{State} & \textbf{Best Val} & \textbf{Best Step} & \textbf{Test} \\
\midrule
\multirow{9}{*}{PaperSearchQA}
& 1 & 4.7 & Failed & --- & --- & --- \\
& 2 & 3.8 & Failed & --- & --- & --- \\
& 3 & 165.7 & Cancelled & 0.384 & 50 & 0.384 \\
& 4 & 137.1 & Cancelled & 0.392 & 80 & 0.366 \\
& 5 & 167.9 & Completed & 0.392 & 120 & 0.394 \\
& 6 & 8.6 & Failed & 0.374 & 200 & 0.386 \\
& 7 & 161.5 & Completed & \textbf{0.396} & 260 & \textbf{0.402} \\
& 8 & 6.0 & Failed & --- & --- & --- \\
& 9 & 53.0 & Cancelled & 0.376 & 160 & 0.394 \\
\midrule
\multirow{6}{*}{WildSci}
& 1 & 136.8 & Cancelled & 0.591 & 20 & 0.561 \\
& 2 & 71.7 & Cancelled & 0.593 & 25 & 0.571 \\
& 3 & 110.8 & Cancelled & 0.619 & 45 & 0.538 \\
& 4 & 111.8 & Cancelled & \textbf{0.621} & 55 & 0.566 \\
& 5 & 17.1 & Completed & 0.597 & 75 & 0.570 \\
& 6 & 39.8 & Cancelled & 0.611 & 75 & \textbf{0.603} \\
\midrule
\multirow{6}{*}{SSMR-Bench}
& 1 & 328.8 & Cancelled & 0.828 & 90 & 0.820 \\
& 2 & 21.2 & Completed & 0.816 & 81 & 0.788 \\
& 3 & 48.1 & Cancelled & \textbf{0.846} & 85 & 0.800 \\
& 4 & 64.2 & Completed & 0.824 & 101 & 0.820 \\
& 5 & 65.4 & Completed & 0.814 & 105 & 0.796 \\
& 6 & 72.2 & Cancelled & 0.814 & 105 & 0.804 \\
\midrule
\multirow{7}{*}{ChemCoT}
& 1 & 399.7 & Cancelled & 0.281 & 125 & 0.376 \\
& 2 & 23.8 & Failed & 0.253 & 120 & 0.371 \\
& 3 & 55.4 & Cancelled & 0.271 & 125 & 0.334 \\
& 4 & 96.6 & Cancelled & \textbf{0.299} & 130 & \textbf{0.380} \\
& 5 & 46.8 & Cancelled & 0.291 & 140 & 0.380 \\
& 6 & 47.3 & Cancelled & 0.295 & 140 & 0.377 \\
& 7 & 39.2 & Cancelled & 0.259 & 125 & 0.356 \\
\bottomrule
\end{tabular}%
}
\end{table*}

\subsection{Per-Node Results: \ours{}}

Tables~\ref{tab:llmzero_ssmr_nodes}--\ref{tab:llmzero_wildsci_nodes} report per-node results for \ours{} on all four tasks.

\begin{table*}[h]
\caption{\ours{} per-node results on SSMR-Bench (10 nodes, 64$\times$A100). Val = aggregate validation score. Test = average of 4 subtask test scores at best validation step. ES = early-stopped.}
\label{tab:llmzero_ssmr_nodes}
\centering
\scalebox{0.82}{%
\small
\begin{tabular}{c*{7}{c}}
\toprule
\textbf{Node} & \textbf{Wall-hrs} & \textbf{GPU-hrs} & \textbf{Status} & \textbf{Best Val} & \textbf{Best Step} & \textbf{Test (avg)} \\
\midrule
0 & 4.5 & 291 & Completed & 0.700 & 30 & 0.658 \\
1 & 2.5 & 162 & ES & 0.672 & 33 & 0.658 \\
2 & 2.5 & 162 & ES & 0.682 & 36 & 0.682 \\
3 & 7.0 & 451 & ES & 0.732 & 66 & 0.706 \\
4 & 13.2 & 846 & Completed & 0.830 & 153 & 0.822 \\
\textbf{5} & \textbf{7.3} & \textbf{467} & \textbf{ES} & \textbf{0.870} & \textbf{180} & \textbf{0.822} \\
6 & 12.8 & 821 & ES & 0.764 & 99 & 0.738 \\
7 & 5.8 & 371 & ES & 0.558 & 48 & 0.526 \\
8 & 2.5 & 162 & ES & 0.858 & 195 & 0.838 \\
9 & 6.7 & 428 & Crashed & 0.856 & 192 & 0.818 \\
\midrule
\multicolumn{2}{l}{\textbf{Total}} & \textbf{4,159} & & & & \\
\bottomrule
\end{tabular}%
}
\end{table*}

\begin{table*}[h]
\caption{\ours{} per-node results on ChemCoT (16 nodes, 64$\times$A100). Test = avg(und\_avg, edit\_avg, rxn\_avg) at best validation step.}
\label{tab:llmzero_chemcot_nodes}
\centering
\scalebox{0.82}{%
\small
\begin{tabular}{c*{7}{c}}
\toprule
\textbf{Node} & \textbf{Wall-hrs} & \textbf{GPU-hrs} & \textbf{Status} & \textbf{Best Val} & \textbf{Best Step} & \textbf{Test (avg)} \\
\midrule
0 & 7.3 & 469 & Completed & 0.292 & 36 & 0.330 \\
1 & 4.6 & 294 & Fix needed & 0.312 & 72 & 0.359 \\
2 & 2.6 & 164 & ES & 0.310 & 78 & 0.336 \\
3 & 4.4 & 284 & Completed & 0.332 & 117 & 0.374 \\
4 & 2.6 & 163 & ES & 0.326 & 120 & 0.368 \\
5 & 2.8 & 180 & ES & 0.328 & 132 & 0.386 \\
6 & 7.6 & 484 & ES & 0.264 & 36 & 0.313 \\
7 & 5.6 & 356 & ES & 0.230 & 27 & 0.224 \\
8 & 6.6 & 422 & ES & 0.246 & 30 & 0.264 \\
9 & 7.2 & 460 & Fix needed & 0.266 & 18 & 0.302 \\
10 & 13.9 & 889 & ES & 0.324 & 60 & 0.384 \\
\textbf{11} & \textbf{15.5} & \textbf{992} & \textbf{Fix needed} & \textbf{0.356} & \textbf{180} & \textbf{0.406} \\
12 & 40.1 & 2,566 & Debug & 0.322 & 96 & 0.356 \\
13 & 8.1 & 516 & ES & 0.322 & 78 & 0.380 \\
14 & 2.6 & 164 & ES & 0.330 & 126 & 0.377 \\
15 & 25.2 & 1,610 & ES & 0.330 & 120 & 0.374 \\
\midrule
\multicolumn{2}{l}{\textbf{Total}} & \textbf{10,013} & & & & \\
\bottomrule
\end{tabular}%
}
\end{table*}

\begin{table*}[h]
\caption{\ours{} per-node results on PaperSearchQA (16 nodes, 32$\times$A100). Test = single-metric test score at best validation step.}
\label{tab:llmzero_psqa_nodes}
\centering
\scalebox{0.82}{%
\small
\begin{tabular}{c*{7}{c}}
\toprule
\textbf{Node} & \textbf{Wall-hrs} & \textbf{GPU-hrs} & \textbf{Status} & \textbf{Best Val} & \textbf{Best Step} & \textbf{Test} \\
\midrule
0 & 19.1 & 611 & Completed & 0.388 & 114 & 0.390 \\
1 & 15.7 & 501 & Completed & 0.408 & 123 & 0.410 \\
2 & 18.8 & 601 & Completed & 0.408 & 141 & 0.422 \\
3 & 4.0 & 129 & ES & 0.396 & 144 & 0.400 \\
4 & 3.5 & 113 & ES & 0.390 & 123 & 0.400 \\
\textbf{5} & \textbf{22.3} & \textbf{714} & \textbf{Completed} & \textbf{0.420} & \textbf{303} & \textbf{0.426} \\
6 & 40.0 & 1,281 & Completed & 0.390 & 246 & 0.378 \\
7 & 5.5 & 177 & ES & 0.402 & 165 & 0.398 \\
8 & 6.0 & 193 & ES & 0.414 & 306 & 0.424 \\
9 & 10.0 & 321 & ES & 0.400 & 141 & 0.396 \\
10 & 6.3 & 201 & ES & 0.402 & 162 & 0.398 \\
11 & 40.0 & 1,281 & Completed & 0.380 & 300 & 0.372 \\
12 & 4.0 & 129 & ES & 0.398 & 147 & 0.400 \\
13 & 3.5 & 113 & ES & 0.396 & 162 & 0.402 \\
14 & 4.0 & 129 & ES & 0.392 & 165 & 0.400 \\
15 & 4.8 & 153 & ES & 0.404 & 171 & 0.406 \\
\midrule
\multicolumn{2}{l}{\textbf{Total}} & \textbf{6,646} & & & & \\
\bottomrule
\end{tabular}%
}
\end{table*}

\begin{table*}[h]
\caption{\ours{} per-node results on WildSci (16 nodes, 32$\times$A100). Test = average of 9 domain test scores at best validation step.}
\label{tab:llmzero_wildsci_nodes}
\centering
\scalebox{0.82}{%
\small
\begin{tabular}{c*{7}{c}}
\toprule
\textbf{Node} & \textbf{Wall-hrs} & \textbf{GPU-hrs} & \textbf{Status} & \textbf{Best Val} & \textbf{Best Step} & \textbf{Test (avg)} \\
\midrule
0 & 7.3 & 234 & Completed & 0.608 & 36 & 0.548 \\
1 & 10.3 & 329 & ES & 0.626 & 51 & 0.548 \\
2 & 16.8 & 537 & Completed & 0.624 & 105 & 0.551 \\
3 & 9.3 & 297 & ES & 0.618 & 54 & 0.565 \\
4 & 14.6 & 466 & ES & 0.632 & 159 & 0.574 \\
5 & 14.5 & 464 & Completed & 0.620 & 96 & 0.593 \\
6 & 30.0 & 960 & Completed & 0.636 & 189 & 0.586 \\
7 & 17.5 & 560 & Completed & 0.628 & 186 & 0.573 \\
8 & 7.8 & 249 & ES & 0.636 & 210 & 0.589 \\
9 & 7.0 & 225 & ES & 0.636 & 216 & 0.573 \\
10 & 14.1 & 450 & ES & 0.630 & 216 & 0.569 \\
11 & 40.0 & 1,281 & Completed & 0.622 & 120 & 0.560 \\
12 & 4.0 & 129 & ES & 0.620 & 195 & 0.588 \\
13 & 9.8 & 313 & ES & 0.622 & 153 & 0.563 \\
14 & 13.8 & 442 & ES & 0.620 & 129 & 0.586 \\
\textbf{15} & \textbf{13.6} & \textbf{436} & \textbf{Completed} & \textbf{0.640} & \textbf{174} & \textbf{0.585} \\
\midrule
\multicolumn{2}{l}{\textbf{Total}} & \textbf{7,370} & & & & \\
\bottomrule
\end{tabular}%
}
\end{table*}

\section{Human Knowledge Injection}
\label{app:knowledge}

Even the latest LLMs frequently misinterpret domain-specific training metrics and hyperparameters in GRPO/PPO. For example, models consistently confuse PPO policy clipping with response length clipping. To ground the LLM's reasoning, we inject structured human-written descriptions for metrics and hyperparameters into the prompt. Importantly, these descriptions do \emph{not} restrict the LLM to only modifying listed parameters; the agent can change any hyperparameter in the training configuration, including those not covered by the guide.

\subsection{Metric Descriptions}
\label{app:metric_desc}

Figure~\ref{fig:metric_desc} lists the 12 metric descriptions injected into both the proposer and early stopper prompts. These address common LLM misinterpretations.

\begin{figure*}[p]
\begin{tcolorbox}[colback=metricbg, colframe=metricframe, title={\textbf{Metric Descriptions (injected into both proposer and early stopper prompts)}}, fonttitle=\small, fontupper=\scriptsize, left=4pt, right=4pt, top=2pt, bottom=2pt]
\begin{tabular}{@{}p{4.2cm}p{10.3cm}@{}}
\texttt{actor/entropy} & Policy entropy measuring exploration. Healthy: dropping with rising rewards. Problem: both dropping (losing generality without gains). \\[2pt]
\texttt{actor/kl\_loss} & KL divergence loss measuring policy divergence from old policy. Should remain small and stable. \\[2pt]
\texttt{actor/ppo\_kl} & PPO KL divergence. \textbf{Critical:} if ppo\_kl spikes AND rewards drop afterward, training has collapsed. \\[2pt]
\texttt{actor/pg\_loss} & Policy gradient loss (primary training signal). Should generally decrease as policy improves. \\[2pt]
\texttt{actor/pg\_clipfrac} & PPO policy clipping fraction (upper). Ratio of updates hitting clip\_ratio\_high. High ($>$0.3) = many updates being prevented. \textbf{Unrelated to response\_length/clip\_ratio.} \\[2pt]
\texttt{actor/pg\_clipfrac\_lower} & PPO policy clipping fraction (lower). Shows how often PPO constrains updates. \textbf{Unrelated to response\_length/clip\_ratio.} \\[2pt]
\texttt{actor/grad\_norm} & Gradient norm magnitude. Very high = instability; very low = vanishing gradients. \\[2pt]
\texttt{critic/rewards/mean} & Mean predicted rewards. Key indicator of training progress. Should generally increase. \\[2pt]
\texttt{response\_length/mean} & Average response length in tokens. Bounded by max\_response\_length. Interpret alongside rewards. \\[2pt]
\texttt{response\_length/clip\_ratio} & Response truncation ratio (responses hitting max length). \textbf{NOT related to PPO clip\_ratio.} High = responses being cut off. \\[2pt]
\texttt{timing\_s/step} & Time per step. Sudden increases may indicate memory issues or inefficient batch sizes. \\[2pt]
\texttt{perf/max\_memory\_reserved\_gb} & Max GPU memory reserved. Monitors OOM risk. Helps optimize batch sizes. \\
\end{tabular}
\end{tcolorbox}
\caption{Human-written metric descriptions injected into agent prompts to ground LLM reasoning about training dynamics.}
\label{fig:metric_desc}
\end{figure*}

\subsection{Hyperparameter Descriptions}
\label{app:hp_desc}

Figure~\ref{fig:hp_desc} provides the hyperparameter reference guide injected into the proposer prompt. Parameters are organized by functional category. The LLM is not restricted to modifying only these parameters; it can change any value in the training configuration. \revised{For readability, some entries in Figure~\ref{fig:hp_desc} are compressed relative to the injected prompt; for example, the \texttt{rollout.n} entry is followed in the actual prompt by a second line, ``Increase for better learning signal; decrease to save compute,'' so that among all tracked hyperparameters only \texttt{kl\_loss\_coef} and \texttt{max\_response\_length} carry a strictly single-directional cue. This is relevant to the directional-prompt discussion in \S\ref{sec:patterns}.}

\begin{figure*}[p]
\begin{tcolorbox}[colback=hpbg, colframe=hpframe, title={\textbf{Hyperparameter Reference Guide (injected into proposer prompt; LLM not restricted to this list)}}, fonttitle=\small, fontupper=\scriptsize, left=4pt, right=4pt, top=2pt, bottom=2pt]
\begin{tabular}{@{}p{4.5cm}p{10.5cm}@{}}
\multicolumn{2}{@{}l}{\textbf{\textit{Learning Rates \& Optimization}}} \\
\texttt{weight\_decay} & L2 regularization. Increase if large train/val gap; decrease if underfitting. \\[1pt]
\texttt{grad\_clip} & Max gradient norm. Decrease if NaN losses or collapse; increase if gradients consistently small. \\[3pt]
\multicolumn{2}{@{}l}{\textbf{\textit{Batch Sizes}}} \\
\texttt{ppo\_micro\_batch\_size\_per\_gpu} & Micro-batch for gradient accumulation. Critical: ppo\_mini\_batch\_size must be divisible by this. \\[3pt]
\multicolumn{2}{@{}l}{\textbf{\textit{PPO/GRPO Clipping}}} \\
\texttt{clip\_ratio\_low} & PPO lower bound. Keep at default; increasing may collapse sampling space. \\[1pt]
\texttt{clip\_ratio\_high} & PPO upper bound. Higher values enhance entropy and diversity. \\[3pt]
\multicolumn{2}{@{}l}{\textbf{\textit{KL Divergence \& Regularization}}} \\
\texttt{kl\_loss\_coef} & KL loss weight constraining policy changes. Only increase if KL spikes coincide with reward collapse. \\[1pt]
\texttt{entropy\_coeff} & Entropy bonus weight. \textbf{Critical: must remain 0 for GRPO.} \\[1pt]
\texttt{norm\_adv\_by\_std\_in\_grpo} & Normalize advantages by std. Generally keep True for stability. \\[3pt]
\multicolumn{2}{@{}l}{\textbf{\textit{Rollout \& Sampling}}} \\
\texttt{rollout.n} & Samples per prompt. Higher improves advantage estimates, increases compute linearly. \\[1pt]
\texttt{temperature} & Sampling temperature. Higher increases exploration and diversity. \\[3pt]
\multicolumn{2}{@{}l}{\textbf{\textit{Sequence \& Memory Management}}} \\
\texttt{max\_response\_length} & Max response tokens. Increase if response\_length/clip\_ratio is high. \\[1pt]
\texttt{max\_num\_seqs} & Max concurrent vLLM sequences. Decrease if OOM; increase for GPU utilization. \\[1pt]
\texttt{tensor\_model\_parallel\_size} & GPUs for model sharding. Use minimum that fits in memory. \\[3pt]
\multicolumn{2}{@{}l}{\textbf{\textit{Training Configuration}}} \\
\texttt{total\_epochs} & Passes through data. Increase if validation still improving at final epoch. \\
\end{tabular}
\end{tcolorbox}
\caption{Human-written hyperparameter descriptions injected into the proposer prompt, organized by functional category.}
\label{fig:hp_desc}
\end{figure*}

\section{Agent Prompts}
\label{app:prompts}

We provide the full proposer and early stopper prompts below. These prompts were minimally tuned: we deliberately avoided hardcoding any numeric thresholds or task-specific values, relying instead on the LLM's reasoning to interpret metrics in context. This design leaves substantial room for improvement through prompt engineering, and we expect that more carefully crafted prompts could further improve search efficiency.

\subsection{Proposer Agent Prompt}
\label{app:proposer_prompt}

The proposer receives the previous training configuration (as YAML), step-level metric summaries (text and/or plots), and produces a diagnosis with hyperparameter suggestions. The template (Figure~\ref{fig:proposer_prompt}) is instantiated with run-specific data at each iteration; placeholders in \texttt{\{braces\}} are filled dynamically.

\begin{figure*}[p]
\begin{tcolorbox}[colback=promptbg, colframe=promptframe, title={\textbf{Proposer Agent Prompt Template}}, fonttitle=\small, left=4pt, right=4pt, top=4pt, bottom=4pt]
\begin{lstlisting}
You are a hyperparameter optimization expert. Your goal is to maximize the validation
score by proposing hyperparameter changes for the next training run.

{previous_training_config}

{early_stop_notice}

### Training Step Metrics Summary
The training ran for {training_steps} steps:

{step_metrics_summary}

{metric_descriptions}

{hyperparameter_guide}

### Metric Interpretation
**PRIMARY:** Rewards (critic/rewards/mean) and validation scores measure overall
performance. Judge success by these metrics.
**DIAGNOSTIC:** actor/entropy, actor/kl_loss, actor/grad_norm, etc. Only relevant
when primary metrics degrade. Rising KL with stable rewards is acceptable.

### Your Task

Follow these steps **in order**. Do NOT suggest `total_epochs` until Step 4.

**Step 1 - Diagnosis:** Summarize the training health. What are the key trends in
primary metrics? Is the run healthy, plateauing, or collapsing? Note any inflection
points or anomalies.

**Step 2 - Hyperparameter suggestions:** Based on your diagnosis, propose specific
hyperparameter changes for the next run. Do NOT include `total_epochs` here
(computed in Step 4).
{suggestion_count_guidance}

**Step 3 - Checkpoint decision:**
{checkpoint_info}

**Step 4 - Epoch computation:** Now that you have decided all hyperparameters
(including `train_batch_size`) and the checkpoint strategy, compute `total_epochs`:
- If resuming: old_steps_per_epoch = total_training_steps / old_total_epochs,
  new_steps_per_epoch = old_steps_per_epoch * old_batch_size / new_batch_size,
  total_epochs = ceil((resume_step + ADDITIONAL_TRAINING_STEPS) / new_steps_per_epoch)
- If training from scratch: choose `total_epochs` based on target training steps
  and new batch size.

Show the math explicitly.
\end{lstlisting}
\end{tcolorbox}
\caption{Proposer agent prompt template. Placeholders are filled with run-specific data at each search iteration.}
\label{fig:proposer_prompt}
\end{figure*}

\subsection{Early Stopper Agent Prompt}
\label{app:early_stopper_prompt}

The early stopper receives comparison plots (current run in blue vs.\ best run in green) and decides whether to continue or stop. The prompt template is shown in Figure~\ref{fig:early_stopper_prompt}.

\begin{figure*}[p]
\begin{tcolorbox}[colback=promptbg, colframe=promptframe, title={\textbf{Early Stopper Agent Prompt Template}}, fonttitle=\small, left=4pt, right=4pt, top=4pt, bottom=4pt]
\begin{lstlisting}
You are an expert at analyzing RL for LLM post training using visual plots. Your task
is to decide if a run should continue or stop early.

### Objective
Your ONLY goal is to maximize the validation score. CONTINUE if the current run's
validation score trajectory has potential to exceed the best validation score seen so
far. STOP only when there is no realistic chance of beating it.

### Best Run Information
{best_run_summary}

### Current Run Progress (at checkpoint)
{current_run_summary}

### Visual Comparison
The plots show training metrics over time. Blue = current run, green dashed = best
run, both may start from an intermediate checkpoint.

Focus on the **validation score plot**. Ask: can the blue curve's peak eventually
exceed the green curve's peak?

{metric_descriptions}

### Decision Rules
**Default action is CONTINUE.** When uncertain, CONTINUE. Only STOP if ALL of these
are true:
1. Current step >= 10 (too early to judge before that)
2. The validation score trajectory has no realistic chance of exceeding the best
   validation score seen so far, considering the improvement rate, not just the
   current value. A run behind the best can still win if its trajectory is steeper;
   a run ahead can still lose if it is plateauing.

**Only the validation score determines STOP/CONTINUE.** All other metrics (rewards,
entropy, KL, gradient norms, clip_ratio, etc.) provide context but NEVER justify
stopping on their own.

**Response Format:**
REASON: [1-2 sentences on whether the validation score trajectory can beat the best
validation score seen so far]
DECISION: [CONTINUE or STOP]
\end{lstlisting}
\end{tcolorbox}
\caption{Early stopper agent prompt template. Invoked every 900 seconds during training.}
\label{fig:early_stopper_prompt}
\end{figure*}

\section{Ethics and Artifact Documentation}
\label{app:ethics}

\paragraph{Potential risks.}
\ours{} automates hyperparameter search for RL post-training, which could lower the barrier to fine-tuning language models for harmful purposes. However, the system requires substantial compute, limiting misuse to well-resourced actors who already have access to equivalent capabilities.

\paragraph{Artifact licenses.}
All artifacts are used under their respective open-source licenses.

\begin{center}
\small
\begin{tabular}{ll}
\toprule
\textbf{Artifact} & \textbf{License} \\
\midrule
Qwen3 (0.6B--8B) & Apache 2.0 \\
ChemCoTBench & CC BY 4.0 \\
PaperSearchQA & MIT \\
SSMR-Bench & MIT \\
WildSci & CC BY 4.0 \\
VeRL & Apache 2.0 \\
Ray & Apache 2.0 \\
\bottomrule
\end{tabular}
\end{center}

\paragraph{Intended use consistency.}
All datasets are used for their intended purpose of evaluating language model capabilities on domain-specific reasoning tasks. VeRL and Ray are used for distributed RL training, consistent with their documented use cases.

\paragraph{Artifact documentation.}
We use 4 evaluation datasets (5{,}000 train / 500 val / 500 test each), Qwen3 models (0.6B--8B) with LoRA fine-tuning, a modified version of VeRL for GRPO training, and Ray for distributed orchestration on EKS clusters with A100 GPUs. All datasets are publicly available. We do not release trained model weights; we release discovered strategy configurations to enable reproduction. \revised{Upon acceptance we will additionally release the search framework code, all discovered strategy configurations, the task reward functions, and the full agent prompts (proposer and early stopper) to support reproducibility. Our new experiments on the public VeRL 0.7.1 release (Appendix~\ref{app:public_stack}) re-confirm the main claims on a fully public stack, independent of the internal VeRL build used for the main results.}

\section{Use of AI}
\label{app:ai_use}

We used AI-based tools to assist with grammar and writing clarity of the paper.

\end{document}